\documentclass[twocolumn]{svjour3}          % twocolumn
\smartqed  % flush right qed marks, e.g. at end of proof
\usepackage{graphicx}
\usepackage{natbib}
\usepackage[english]{babel}
\usepackage{hyphenat}
\usepackage[colorlinks,citecolor=blue]{hyperref}
\journalname{Biological Cybernetics}
\begin{document}

\title{Short\hyp term plasticity as cause-effect hypothesis testing in distal reward learning %\thanks{Grants or other notes
}


\author{Andrea Soltoggio%\and
        %Second Author %etc.
}


\institute{Andrea Soltoggio \at
              Computer Science Department\\
              Loughborough University,
              LE11 3TU, Loughborough, UK.\\
              \email{a.soltoggio@lboro.ac.uk}           %  \\
}

\date{Received: 16 December 2013 / Accepted: 6 August 2014}

\maketitle

\begin{abstract}
Asynchrony, overlaps and delays in sensory\hyp motor signals introduce ambiguity as to which stimuli, actions, and rewards are causally related. Only the repetition of reward episodes helps distinguish true cause\hyp effect relationships from coincidental occurrences. In the model proposed here, a novel plasticity rule employs short and long\hyp term changes to evaluate hypotheses on cause\hyp effect relationships. Transient weights represent hypotheses that are consolidated in long\hyp term memory only when they consistently predict or cause future rewards. The main objective of the model is to preserve existing network topologies when learning with ambiguous information flows. Learning is also improved by biasing the exploration of the stimulus\hyp response space towards actions that in the past occurred before rewards. The model indicates under which conditions beliefs can be consolidated in long\hyp term memory, it suggests a solution to the plasticity\hyp stability dilemma, and proposes an interpretation of the role of short\hyp term plasticity.
\end{abstract}

\section{Introduction}
\label{intro}
Living organisms endowed with a neural system constantly receive sensory information and perform actions. Occasionally, actions lead to rewards or punishments in the near future, e.g.\,tasting food after following a scent \citep{staubli1987}. The exploration of the stimulus\hyp action patterns, and the exploitation of those patterns that lead to rewards, was observed in animal behavior and named \emph{operant conditioning} \citep{thorndike1911,skinner1953}. Mathematical abstractions of operant conditioning are formalized in algorithms that maximize a reward function in the field of reinforcement learning \citep{suttonBarto1998}. The maximization of reward functions was also implemented in a variety of neural network models \citep{linPhDThesis1993,pennartz1997,schultzDayanMontague1997,bosmanLeeuwenWemmenhove2004,XieSeung2004,florian2007,farriesFairhall2007,barasMeir2007,legensteinChaseSchwartzMaass2010,fremauxSprekelerGerstner2010,friedrichUrbanczikSenn2010NECO}, and is inspired and justified by solid biological evidence on the role of neuromodulation in associative and reward learning \citep{wiseRompre1989,schultzApicellaLjungberg1993,swartzentruber1995modulatory,pennartz1996,schultz1998,nitzKargoFleisher2007,berridge2007,redgraveGurneyReynolds2008}. The utility of modulatory dynamics in models of reward learning and behavior is also validated by closed-loop robotic neural controllers \citep{ziemkeThieme2002,spornsAlexander2002,alexanderSporns2002,spornsAlexander2003,soltoggioBullinariaMattiussiDuerrFloreano2008,coxJrichmar2009}.

Neural models encounter difficulties when delays occur between perception, actions, and rewards. A first issue is that a neural network needs a memory, or a trace, of previous events in order to associate them to later rewards. But a second even trickier problem lies in the environment: if there is a continuous flow of stimuli and actions,  unrelated stimuli and actions intervene between causes and rewards. The environment is thus ambiguous as to which stimulus\hyp action pairs lead to a later reward. Concomitant stimuli and actions also introduce ambiguity. In other words, any learning algorithm faces a condition in which one single reward episode does not suffice to understand which of the many preceding stimuli and actions are responsible for the delivery of the reward.   %This problem is faced by a neural system that is continuously fed with inputs and performs actions and infers which neural activity in the past is responsible for the delivery of a reward in the present. 
This problem was called the \emph{distal reward problem} \citep{hull1943}, or \emph{credit assignment problem} \citep{sutton1984phd,suttonBarto1998}. Credit assignment is a general machine learning problem. Neural  models that solve it may help clarify which computation is employed by animals to deal with asynchronous and deceiving information. Learning in ambiguous conditions is in fact an ubiquitous type of neural learning observed in mammals as well as in simpler neural systems \citep{brembs2003operant} as that of the invertebrate Aplysia \citep{brembsLorenzettiReyesBaxterByrne2002} or the honey bee \citep{hammer1995learning,menzelMueller1996,gilDeMarcoMenzel2007}. 

When environments are ambiguous due to delayed rewards, or due to concomitant stimuli and actions, the only possibility of finding true cause\hyp effect relationships is to observe repeated occurrences of a reward. By doing that, it is possible to assess the probability of certain stimuli and actions to be the cause of the observed reward. Previous neural models, e.g.\,\citep{izhikevich2007,fremauxSprekelerGerstner2010,friedrichUrbanczikSenn2011,soltoggioSteilNeuralComputation2013}, solve the distal reward problem applying small weight changes whenever an event indicates an increased or decreased probability of particular pathways to be associated with a reward. With a sufficiently low learning rate, and after repeated reward episodes, the reward\hyp inducing synapses grow large, while all other synapses sometimes increase and sometimes decrease their weights. Those approaches may perform well in reward maximization tasks, but they also cause deterioration of synaptic values because the whole modulated network constantly undergoes synaptic changes across non\hyp reward\hyp inducing synapses. %Even algorithms that are not affected by the so called ``unsupervised bias'', are destructive of network topologies because the ambiguity introduced by delays causes weight updates of synapses that are not related to rewards.
For this reason, only limited information, i.e.\,those stimulus\hyp action pairs that are frequently rewarded, can be retained even in large networks because the connectivity is constantly rewritten. Interestingly, the degradation of synapses occurs also as a consequence of spontaneous activity as described in \cite{fusi2005cascade}. In general, continuous learning, or synapses that are always plastic, pose a treat to previously acquired memory \citep{senn2005learning,fusi2006eluding,leibold2008sparseness}. Delayed rewards worsen the problem because they amplify synaptic changes caused by reward-unrelated activity. While learning with delayed rewards, current models suffer particularly from the so called \emph{plasticity}\hyp \emph{stability} dilemma, and \emph{catastrophic} \emph{forgetting} \citep{grossberg1988,robins1995,abrahamRobins2005}. 
 
Synapses may be either coincidentally or causally active before reward deliveries, but which of the two cases applies is unknown due to the ambiguity introduced by delays. How can a system solve this apparent dilemma, and correctly update reward\hyp inducing weights and leaving the others unchanged? The novel idea in this study is a distinction between two components of a synaptic weight---a volatile component and a consolidated component. Such as distinction is not new in connectionist models \citep{hinton1987using,schmidhuber1992,levy1995connectionist,tieleman2009using,bullinaria2009}, however, in the proposed study the idea is extended to model hypothesis testing and memory consolidation with distal rewards. The volatile (or transient) component of the weight may increase or decrease at each reward delivery without immediately affecting the long\hyp term component. It decays over time, and for this reason may be seen as a particular form of short\hyp term plasticity. In the rest of the paper, the terms \emph{volatile}, \emph{transient} and \emph{short\hyp term} are used as synonyms to indicate the component of the weight that decays over time. In contrast, \emph{consolidated}, \emph{long-term}, or \emph{stable} are adjectives used to refer to the component of the weight that does not decay over time. %STP is employed as an exploratory tool that tests hypotheses of cause\hyp effect relationships.  %The proposed algorithm modifies the dynamics of the plasticity in \cite{izhikevich2007,soltoggioSteilNeuralComputation2013} to model explicitly a process that tests hypotheses of cause-effect relationships. %short\hyp term plasticity is used in this model to explore and test cause-effect hypotheses. 

Short\hyp term volatile weights are \emph{hypotheses} of how likely stimulus\hyp action pairs lead to future rewards. If not confirmed by repeated disambiguating instances, short\hyp term weights decay without affecting the long\hyp term configuration of the network. Short\hyp term synaptic weights and the plasticity that regulates them can be interpreted as implementing Bayesian belief \citep{howsonUrbach1989}, and the proposed model interpreted as a special case of a learning Bayesian network \citep{heckermanGeigerHaussler1995,ben-gal2007}. Short\hyp term weights that grow large are therefore those that consistently trigger a reward. The idea in this study is to perform a parsimonious consolidation of weights that have grown large due to repeated and consistent reward\hyp driven potentiation. Such dynamics lead to a consolidation of weights representing established hypotheses.  

The novelty of the model consists in implementing dynamics to test \emph{temporal causal hypotheses} with a transient component of the synaptic weight. Transient weights are increased when the evidence suggests an increased probability of being associated with a future reward. As opposed to \cite{izhikevich2007}, in which a baseline modulation results in a weak Hebbian plasticity in absence of reward, in the current model an anti-Hebbian mechanism leads transient weights to be depressed when the evidence suggests no causal relations to future rewards. The distinction between short and long\hyp term components of a weight allows for an implicit estimation of the probability of a weight to be associated with a reward without changing its long-term consolidated value. When coincidental firing leads to an association, which is however not followed by validating future episodes, long-term weight components remain unchanged. The novel plasticity suggests a nonlinear mechanism of consolidation of a hypothesis in established knowledge during distal reward learning.  Thus, the proposed plasticity rule is named Hypothesis Testing Plasticity (HTP). 

The current model uses eligibility traces with a decay in the order of seconds to bridge stimuli, actions, and rewards. As it will be clarified later, the decay of transient weights acts instead in the order of hours, thereby representing the forgetting of coincidental event\hyp reward sequences that are not confirmed by consistent occurrences. It is important to note that HTP does not replace previous plasticity models of reward learning, it rather complements them with the additional idea of decomposing the weight in two components, one for hypothesis testing, and one for long\hyp term storage of established associations.

In short, HTP enacts two main principles. The first is monitoring correlations by means of short\hyp term weights and actively pursuing exploration of probably rewarding stimulus\hyp action pairs; the monitoring (or hypothesis evaluation) is done without affecting the long\hyp term state of the network. The second principle is that of selecting few established relationships to be consolidated in long\hyp term stable memory. 

HTP is a meta\hyp plasticity scheme and is general to both spiking and rate-based codes. The rule expresses a new theory to cope with multiple rewards, to learn faster and preserve memories of one task in the long term also while learning or performing in other tasks.

%This paper illustrates next the learning problem, describes previous methods and introduces the plasticity model. Simulations demonstrate the properties of the new plasticity rule in comparison to a simpler reward-modulated rule with eligibility traces. The significance and implications of the findings are then discussed.

\section{Method}

This section describes the learning problem, overviews existing plasticity models that solve the distal reward problem, and introduces the novel meta\hyp plasticity rule called Hypothesis Testing Plasticity (HTP).

\subsection{Operant learning with asynchronous and distal rewards}

A newly born learning agent, when it starts to experience a flow of stimuli and to perform actions, has no knowledge of the meaning of inputs, nor of the consequences of actions. The learning process considered here aims at understanding what reward relationships exist between stimuli and actions. 

\begin{figure}
\begin{centering}
\includegraphics[width = 0.4\textwidth]{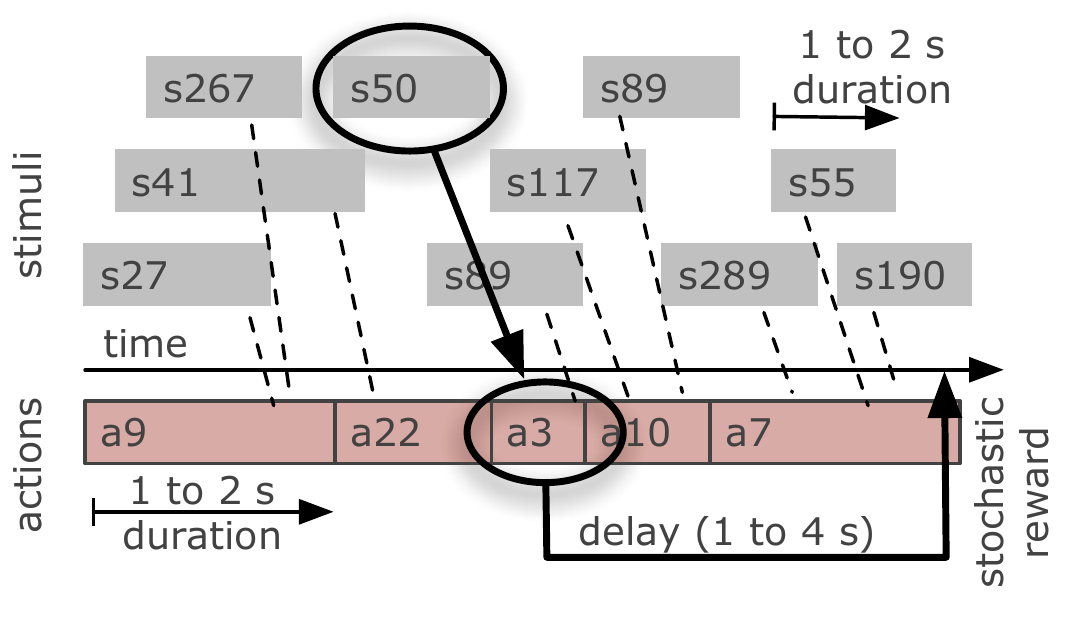}
\caption{Graphical representation of the asynchronous flow of stimuli and actions with delayed rewards. The agent perceives an input flow and performs actions. The delayed rewards, as well as the simultaneous presence of stimuli and actions, cause ambiguity as to which stimulus\hyp action pair is the real cause of a reward. The occurrence of more rewards is the only possibility to disambiguate those relationships and discover the invariant causes. In this graphical illustration, when the reward is delivered, 10 stimulus\hyp action pairs (indicated with dash lines) were active in the recent past. All those pairs may be potentially the cause of the reward: in effect, it is only the pair s50\hyp a3 that caused the reward.}
\label{fig.f0}
\end{centering}
\end{figure}

The overlapping of stimuli and actions represents the coexistence of a flow of stimuli with a flow of actions. Stimuli and actions are asynchronous and initially unrelated. The execution of actions is initially driven by internal dynamics, e.g.\,driven by noise, because the agent's knowledge is a tabula rasa, i.e.\,is unbiased and agnostic of the world. Spontaneous action generation is a form of exploration. A graphical representation of the input\hyp output flow is given in Fig.\,\ref{fig.f0}. 

In the setup of the current experiments, at any moment there might be between zero and three stimuli. Stimuli and actions have a random duration between $1$ and $2$ s. Some actions, if performed when particular stimuli are present, cause the delivery of a global noisy signal later in time (between 1 and 4 s later), which can be seen as a reward, or simply as an unconditioned stimulus. The global reward signal is highly stochastic in the sense that both the delay and the intensity are variable. In the present setting, 300 different stimuli may be perceived at random times. The agent can perform 30 different actions, and the total number of stimulus\hyp action pairs is 9000. The task is to learn which action to perform when particular stimuli are present to obtain a reward.

It is important to note that the ambiguity as to which pairs cause a reward emerges from both the simultaneous presence of more stimuli, and from the delay of a following reward. From a qualitative point of view, whether distinct stimulus\hyp action pairs occurred simultaneously or in sequence has no consequence: a learning mechanism must take into consideration that a set of pairs were active in the recent past. Accordingly, the word \emph{ambiguity} in this study refers to the fact that, at the moment of a reward delivery, several stimulus\hyp action pairs were active in the recent past, and all of them may potentially be the cause of the reward. %More analytical definitions of \emph{ambiguity} could be introduced by defining a measure capturing how many stimulus\hyp action pairs are considered as potential cause of the reward, and therefore eligible for credit assignment.

\subsection{Previous models with synaptic eligibility traces}

In simple neural models, the neural activity that triggers an action, either randomly or elicited by a particular stimulus, is gone when a reward is delivered seconds later. For this reason, standard modulated plasticity rules, e.g.\, \citep{montagueDayanPersonSejnowski1995,soltoggioStanley2012}, fail unless reward is simultaneous with the stimuli. If the reward is not simultaneous with its causes, \emph{eligibility traces} or \emph{synaptic tags} have been proposed as means to bridge the temporal gap \citep{freyMorris1997,wangDenkHaeusser2000,sarkisovWang2008,paepperKempterLeibold2011}. 

Previous models with reward\hyp modulated Hebbian plasticity and eligibility traces were shown to associate past events with following rewards, both in spiking models with spike\hyp timing\hyp dependent plasticity (STDP) \linebreak \citep{izhikevich2007} and in rate\hyp based models with Rarely Correlating Hebbian Plasticity (RCHP) \citep{soltoggioSteilNeuralComputation2013,soltoggioLemmeReinhartSteilFrontiers2013}. RCHP is a filtered Hebbian rule that detects only highly correlating and highly decorrelating activity by means of two thresholds (see Appendix 2): the effect is that of representing sparse (or rare) spiking coincidence also in rate\hyp based models. RCHP was shown in \cite{soltoggioSteilNeuralComputation2013} to have computationally equivalent learning to the  spiking rule (R-STDP) in \cite{izhikevich2007}. 

Spike coincidence in \cite{izhikevich2007}, or highly correlating activity in \cite{soltoggioSteilNeuralComputation2013}, increase synapse\hyp specific eligibility traces. Even with fast network activity (in the millisecond time scale), eligibility traces can last several seconds: when a reward occurs seconds later, it multiplies those traces and reinforces synapses that were active in a recent time window. %A modulatory signal $m$, governed by a fast decay and by the exogenous input reward $r(t)$, converts eligibility traces to weight changes. 
Given a presynaptic neuron $j$ and a postsynaptic neuron $i$, the changes of weights $w_{ji}$, modulation $m$, and  eligibility traces $E_{ji}$, are governed by\begin{eqnarray}
\dot{m}(t) &=& -m(t)/\tau_{m} + \lambda \cdot r(t) + b\label{eq.modulation}\\
\dot{w}_{ji}(t) &=& m(t) \cdot E_{ji}(t)\label{eq.wUpdate}\quad,\label{eq.m}\\
\dot{E}_{ji} &=& -E_{ji}/\tau_{E} + \Theta_{ji}(t)\label{eq.eTupdate}
\end{eqnarray} 
where the modulatory signal $m(t)$ is a leaky integrator of the global reward signal $r(t)$ with a bias $b$; $\tau_{E}$ and $\tau_{m}$ are the time constants of the eligibility traces and modulatory signal; $\lambda$ is a learning rate. The signal $r(t)$ is the reward determined by the environment. The modulatory signal $m(t)$, loosely representing dopaminergic activity, decays relatively quickly with a time constant $\tau_m = 0.1$ s as measured in \cite{wightmanZimmerman1990, garris1994}. In effect, Eq.\,(\ref{eq.modulation}) is a rapidly decaying leaky integrator of instantaneous reward signals received from the environment. The synaptic trace $E$ is a leaky integrator of correlation episodes $\Theta$. In \cite{izhikevich2007}, $\Theta$ is the STDP(t) function; in \cite{soltoggioSteilNeuralComputation2013}, $\Theta$ is implemented by the rate-based Rarely Correlating Hebbian Plasticity (RCHP) that was shown to lead to the same neural learning dynamics of the spiking model in \cite{izhikevich2007}. RCHP is a thresholded Hebbian rule expressed as
\begin{equation}
\Theta_{ji} = \mathrm{RCHP}_{ji}(t) = 
\left\{
\begin{array}{ll}
+\alpha \, & \mathrm{if}\,  v_j(t-t_{pt}) \cdot v_i(t) > \theta_{hi}\\
-\beta \,&\textrm{if}\,  v_j(t-t_{pt}) \cdot v_i(t) < \theta_{lo}\\
0 \, & \textrm{otherwise}\\
\end{array} \right.
\\\label{eq.RCHP}
\end{equation}
where $\alpha$ and $\beta$ are two positive learning rates for correlating and decorrelating synapses respectively, $v(t)$ is the neural output, $t_{pt}$ is the propagation time of the signal from the presynaptic to the postsynaptic neuron, and $\theta_{hi}$ and $\theta_{lo}$ are the thresholds that detect highly correlating and highly decorrelating activities. RCHP is a nonlinear filter on the basic Hebbian rule that ignores most correlations. Note that the propagation time $t_{pt}$ in the Hebbian term implies that the product is not between simultaneous presynaptic and postsynaptic activity, but between presynaptic activity and postsynaptic activity when the signal has reached the postsynaptic neuron. This type of computation attempts to capture the effect of a presynaptic neuron on the postsynaptic neuron, i.e.\,the causal pre-before-post situation \citep{gerstner2010FromHebb}, considered to be the link between the Hebb's postulate and STDP \citep{kempterGerstnerVanHemmen1999}. The regulation of the adaptive threshold is described in the Appendix 2s. A baseline modulation $b$ can be set to a small value and has the function of maintaining a small level of plasticity. 

The idea behind RCHP, which reproduces with rate-based models the dynamics of R-STDP, is that eligibility traces must be created parsimoniously (with rare correlations). When this criterion is respected, both spiking and rate-based models display similar learning dynamics. 

In the current model, the neural state $u_i$ and output $v_{i}$ of a neuron $i$ are computed with a standard rate-based model expressed by
\begin{equation}
u_i(t) = \sum_{j}{(w_{ji} \cdot v_j(t)) + I_{i}}
\label{eq.neuronU}
\end{equation}
\begin{equation}
v_i(t+\Delta t) = 
\left\{
\begin{array}{ll}
\mathrm{tanh}\big(\gamma \cdot u_i(t)\big) + \xi_{i}(t)& \,\, \mathrm{if} \,\,u_i \ge 0\\
\xi_{i}(t)&\,\, \mathrm{if} \,\,u_i < 0
\end{array} \right.
\label{eq.neuron}
\end{equation}
where $w_{ji}$ is the connection weight from a presynaptic neuron $j$ to a postsynaptic neuron $i$; $\gamma$ is a gain parameter set to 0.5; $\xi_{i}(t)$ is a Gaussian noise source with standard deviation 0.02. The input current I is set to 10 when an input is delivered to a neuron. The sampling time is set to $100$ ms, which is also assumed to be the propagation time $t_{pt}$ (Eq.\,(\ref{eq.RCHP})) of signals among neurons. 

\subsection{Hypothesis Testing Plasticity (HTP)}

The dynamics of Eqs.\,(\ref{eq.m}\hyp \ref{eq.eTupdate}) erode existing synapses because the spontaneous network activity during reward episodes causes synaptic correlations and weight\linebreak changes. The deterioration is not only caused by endogenous network activity, but it is also caused by the ambiguous information flow (Fig.\,\ref{fig.f0}). In fact, many \linebreak synapses are often increased or decreased because the corresponding stimulus-action pair is coincidentally active shortly before a reward delivery. Therefore, even if the network was internally silent, i.e.\,there was no spontaneous activity, the continuous flow of inputs and outputs generates correlations that are transformed in weight changes when rewards occur. Such changes are important because they test hypotheses. Unfortunately, if applied directly to the weights, they will eventually wear out existing topologies. 

To avoid this problem, the algorithm proposed in this study explicitly assigns the fluctuating dynamics of Eq.\,(\ref{eq.m}) to a transient  component of the weight. As opposed to the long\hyp term component, the transient component decays over time. Assume, e.g., that one particular synapse had pre and postsynaptic correlating activity just before a reward delivery, but it is not known whether there is a causal relation to the delivery of such a reward, or whether such a correlation was only coincidental. Eq.\,(\ref{eq.m}) increases correctly the weight of that synapse because there is no way at this stage to know whether the relation is causal or coincidental. In the variation proposed here, such a weight increase has a short\hyp term nature because it does not represent the acquisition of established knowledge, but it rather represents the increase of probability that such a synapse is related to a reward delivery. Accordingly, weight changes in Eq.\,(\ref{eq.wUpdate}) are newly interpreted as changes with short\hyp term dynamics \begin{equation}
\dot{w}^{st}_{ji}(t) = -w^{st}_{ji}/\tau_{st} + m(t) \cdot E_{ji}(t)\label{eq.wUpdateNew}\\
\end{equation}
where $w^{st}$ is now a transient component of the weight, and $\tau_{st}$ is the corresponding decay time constant. The time constant of short\hyp term memory $\tau_{st}$ is set to 8 h. In biological studies, short-term plasticity is considered only for potentiation lasting up to $10$ minutes \citep{zucker1989short,fisher1997multiple,zucker2002short}. However, in the idea of this study, the duration of volatile weights represents the duration of an \emph{hypothesis} rather than a specific biological decay. Thus, the value of $\tau_{st}$ can be chosen in a large range. A brief time constant ensures that weights decay quickly if rewards are not delivered. This helps maintain low weights but, if rewards are sparse in time, hypotheses are forgotten too quickly. With sporadic rewards, a longer decay may help preserve hypotheses longer in time. If $\tau_{st}$ is set to large values, hypotheses remain valid for an arbitrary long time. This point indicates that, in the current model, short\hyp term weights are intended primarily as \emph{probabilities} of relationships to be true, rather than simply short \emph{time spans} of certain information.

If a stimulus\hyp action pair is active at a particular point in time, but no reward follows within a given interval (1 to 4 s), it would make sense to infer that such a stimulus\hyp action pair is unlikely to cause a reward. This idea is implemented in HTP by setting the baseline modulation value $b$ in Eq.\,(\ref{eq.m}) to a small negative value. The effect is that of establishing weak anti-Hebbian dynamics across the network in absence of rewards. Such a setting is in contrast to \cite{izhikevich2007} in which the baseline modulation is positive. By introducing a small negative baseline modulation, the activation of a stimulus-action pair, and the consequent increase of $E$, results in a net weight decrement if no reward follows. In other words, high eligibility traces that are not followed by a reward cause a small weight decrease. This modification that decreases a weight if reward does not follow is a core principle in the hypothesis testing mechanism introduced by HTP. By introducing this idea, weights do not need to be randomly depressed by decorrelations, which therefore are not included in the current model. 

Finally, the principles of HTP illustrated above can be applied to a reward\hyp modulated plasticity rule such as R-STDP, RCHP, or any rule capable of computing sparse correlations $\Theta$ in the neural activity, and consequently $E$, in Eq.\,\ref{eq.eTupdate}. In the current study, a rate\hyp based model plus RCHP are employed. In particular, a simplified version of the RCHP, without decorrelations, is expressed as
\begin{equation} 
\Theta_{ji} = \mathrm{RCHP}^{+}_{ji}(t) = 
+1 \quad \mathrm{if}\,  v_j(t-t_{pt}) \cdot v_i(t) > \theta_{hi}
\\\label{eq.RCHP*}
\end{equation}
and 0 otherwise (compare with Eq.\,(\ref{eq.RCHP})). Decorrelations may be nevertheless modelled to introduce weight competition\footnote{In that case, is essential that the traces $E$ are bound to positive values: negative traces that multiply with the negative baseline modulation would lead to unwanted weight increase.}. %In short, R-STDP and RCHP use decorrelations to keep weight low (clearly stated also in \cite{izhikevich2007}), however, they do that by depressing random weights, thereby introducing deteriorating dynamics for existing topologies. Instead, the new approach decreases weights when neurons correlate but no reward follow. The new method is a more logic procedure to update weights in distal reward learning, and its benefits become clear in the simulations.

The overall HTP synaptic weight W is the sum of the short-term and long-term components
\begin{equation}
W_{ji}(t) = w^{st}_{ji}(t) + w^{lt}_{ji}(t)\quad.\label{eq.W}
\end{equation}
As the transient component is also contributing to the overall weight, short\hyp term changes also influences how presynaptic neurons affect postsynaptic neurons, thereby biasing exploration policies as it will be explained in the result section. 

The proposed model consolidates transient weights in long\hyp term weights when the transient values grow large. Such a growth indicates a high probability that the activity across that synapse is involved in triggering following rewards. In other words, after sufficient trials have disambiguated the uncertainty introduced by the delayed rewards, a nonlinear mechanism convert hypotheses to certainties. Previous models \citep{izhikevich2007,obrienSrinivasa2013NC,soltoggioSteilNeuralComputation2013} show a \emph{separation} the of weight values between reward\hyp inducing synapses (high values) and other synapses (low values). In the current model, such a separation is exploited and identified by a threshold $\Psi$ loosely set to a high value, in this particular setting to 0.95 (with weights ranging in [0, 1]).  The conversion is formally expressed as
\begin{equation}
\dot{w}^{lt}_{ji}(t) = \rho \cdot H(w^{st}_{ji}(t) - \Psi)\quad,\label{eq.ltwNewUpdate}
\end{equation}
where $H$ is the Heaviside function and $\rho$ is a consolidation rate, here set to 1/1800 s. Note that in this formulation, $\dot{w}^{lt}_{ji}(t)$ can only be positive, i.e.\,long\hyp term weights can only increase: a variation of the model is discussed later and proposed in the Appendix 1. The consolidation rate $\rho$ means that short\hyp term components are consolidated in long\hyp term components in half an hour when they are larger than the threshold $\Psi$. A one\hyp step instantaneous consolidation (less biologically plausible) was also tested and gave similar results, indicating that the consolidation rate is not crucial. 

The threshold $\Psi$ represents the point at which an hypothesis is considered true, and therefore consolidated in long\hyp term weight. The idea is that, if a particular stimulus\hyp action pair has been active many times consistently before a reward, such stimulus\hyp action pair is indeed causing the reward. Interestingly, because the learning problem is inductive and processes are stochastic, certainty can never be reached from a purely theoretical view point. Assume for example that, on average, every second one reward episode occurs with probability $p = 10^{-2}$ and leads short\hyp term weights that were active shortly before the delivery to grow of 0.05\footnote{The exact increment depends on the learning rate, on the exact circumstantial delay between activity and reward, and on the intensity of the stochastic reward.}. To grow to saturation, a null weight needs 1) to be active approximately $20$ times before reward deliveries and 2) not to be active when rewards are not delivered. If a synapse is not involved in reward delivery, the probability of such a synapse to reach $\Psi$ might be very low in the oder of $p^{20}$, i.e.\,$10^{-40}$. The complex and non\hyp stationary nature of the problem does not allow for a precise mathematical derivation. Such a probability is in fact affected by a variety of environmental and network factors such as the frequency and amount of reward, the total number of stimulus\hyp action pairs, the firing rate of a given connection, the number of intervening events between cause and effect (reward), and the contribution of the weight itself to a more frequent firing. Nevertheless, previous mathematical and neural models that solve the distal reward problem rely on the fact that consistent relationships occurs indeed consistently and more frequently than random events. As a consequence, after a number of reward episodes, the weight that is the true cause of reward has been \emph{accredited} (increased) more than any other weight. The emergence of a \emph{separation} between reward\hyp inducing weights and other weights is observed in \cite{izhikevich2007,obrienSrinivasa2013NC,soltoggioSteilNeuralComputation2013}. %By means of $\Psi$, the current model exploits such a separation and performs a nonlinear transformation that consolidates true relationships and rejects coincidental events. 
The proposed rule exploits this separation between reward\hyp inducing and non\hyp reward\hyp inducing synapses to consolidate established relationship in long\hyp term memory. The dynamics of Eqs.\, (\ref{eq.wUpdateNew}-\ref{eq.ltwNewUpdate}) are referred to as Hypothesis Testing Plasticity (HTP). 

The long-term component, once is consolidated, cannot be undone in the present model. However, reversal learning can be easily implemented by adding complementary dynamics that undo long-term weights if short\hyp term weights become heavily depressed. Such an extension is proposed in the Appendix 1. 

The role of short\hyp term plasticity in improving reward\hyp modulated STDP is also analyzed in a recent study by  \cite{obrienSrinivasa2013NC}. With respect to \linebreak \cite{obrienSrinivasa2013NC}, the idea in the current model is general both to spiking and rate-based coding and is intended to suggest a role of short-term plasticity rather than to model precise biological dynamics. Moreover, it does not employ reward predictors, it focuses on the functional roles of long\hyp term and short\hyp term plasticity, and does not necessitate the Attenuated Reward Gating (ARG). 

Building on models such as  \cite{izhikevich2007,florian2007,friedrichUrbanczikSenn2011,soltoggioSteilNeuralComputation2013}, the current model introduces the concept of testing hypotheses with ambiguous information flow. The novel meta\hyp plasticity model illustrates how the careful promotion of weights to a long\hyp term state allows for retention of memory also while learning new tasks. 

\subsection{Action selection}

Action selection is performed by initiating the action corresponding to the output neuron with the highest activity. Initially, selection is mainly driven by neural noise, but as weights increase, the synaptic strengths bias action selection towards output neurons with strong incoming connections. One action has a random duration between 1 and 2 s. During this time, the action feeds back to the output neuron a signal $I = 0.5$. Such a signal is important to make the \emph{winning} output neuron ``aware'' that it has triggered an action. Computationally, the feedback to the output neuron increases its activity, thereby inducing correlations on that particular input\hyp output pair, and causing the creation of a trace on that particular synapse. Feedback signals to output neurons are demonstrated to help learning also in \cite{urbanczikSenn2009,soltoggioLemmeReinhartSteilFrontiers2013}. The overall structure of the network is graphically represented in Fig.\,\ref{fig.net}. 

Further implementation details are in the Appendix. The Matlab code used to produce the results is made available as support material.
\begin{figure}
\begin{centering}
\includegraphics[width = 0.4\textwidth]{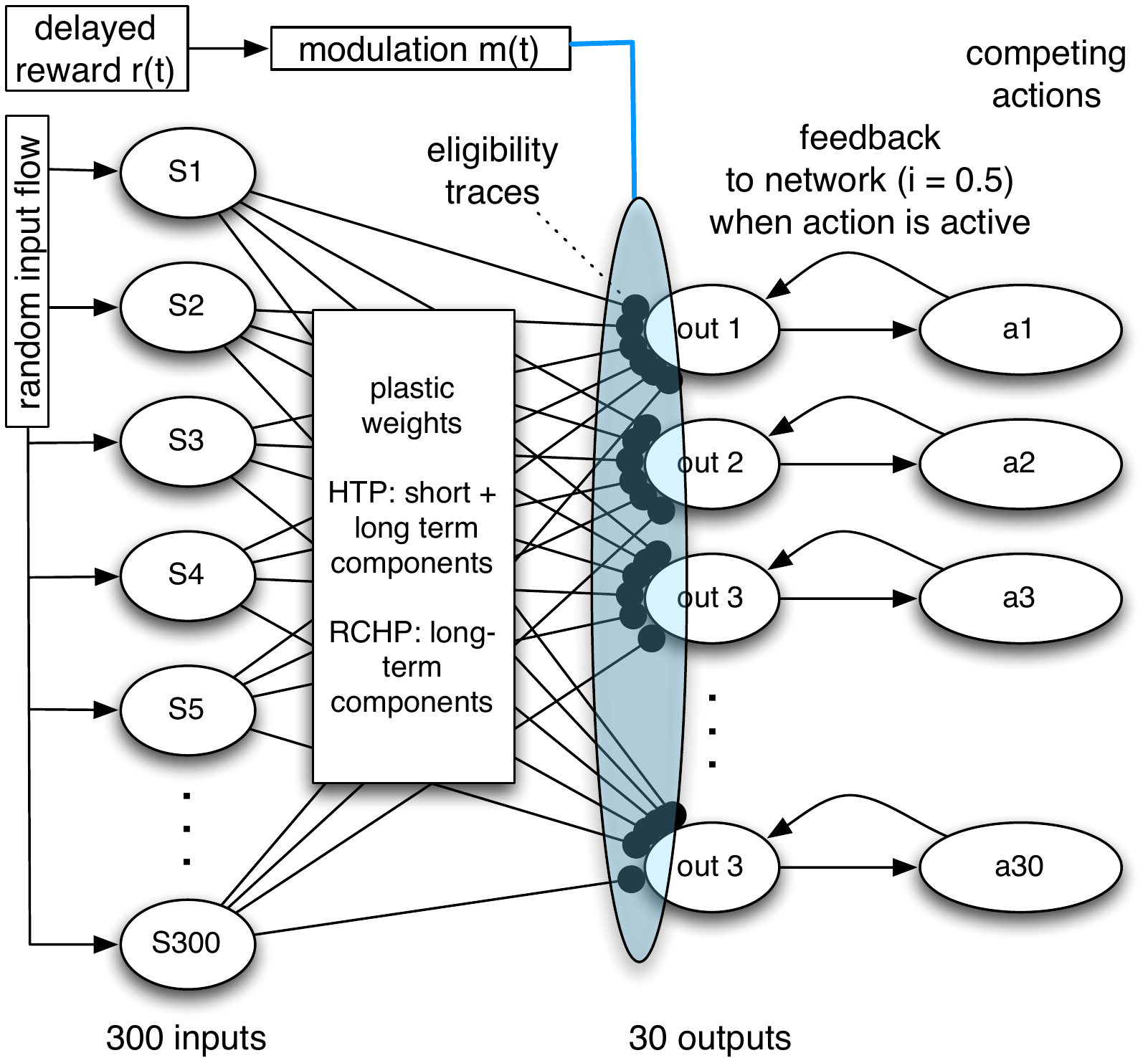}
\caption{Graphical representation of the feed-forward neural network for distal reward learning tested with both the basic RCHP and the novel HTP. One stimulus is delivered to the network by activating a corresponding input neuron. Each weight is plastic and has a trace associated. The modulatory signal is an additional input that modulates the plasticity of all the weights. The sampling time is 100 ms, but the longer temporal dynamics given by delayed rewards is captured by the 4 s time constant of the eligibility traces. The output neuron with the highest activity initiates an action. The action then feeds back to that neuron a feedback signal which helps input and output correlate correctly (see Appendix).}
\label{fig.net}
\end{centering}
\end{figure}

\section{Results}

In this section, simulation results present the computational properties of HTP. A first test is a computational assessment of the extent of weight unwanted change due to distal rewards when one single weight component is used. The learning and memory dynamics of the novel plasticity are tested with the network of Fig.\,\ref{fig.net} on a set of learning scenarios. The dynamics of HTP are illustrated in comparison to those of the single weight component implemented by the basic RCHP.

\subsection{Weight deterioration and stochasticity with distal rewards}
%\footnotesize Note: This section was introduced to satisfy a reviewer: it can be safely skipped without compromising the understanding of the rest of the paper.\vspace{0.3cm}

\normalsize
Algorithms that solve the distal reward problem have so far focused on reward maximization \citep{urbanczikSenn2009,fremauxSprekelerGerstner2010,friedrichUrbanczikSenn2011}. Little attention was given to non\hyp reward\hyp inducing weights. However, non\hyp reward\hyp inducing weights are often the large majority of weights in a network. Their changes are relevant to understand how the whole network evolves over time, and how memory is preserved \citep{senn2005learning}. The test in this section analyzes the side effects of distal rewards on non\hyp reward\hyp inducing weights. 

Assume that a correlating event between two neurons across one synapse $\sigma$ represents a stimulus\hyp action pair that is \emph{not} causing a reward. Due to distal rewards, the synapse $\sigma$ might occasionally register correlation episodes in the time between the real cause and a delayed reward: that is in the nature of the distal reward problem. All synapses that were active shortly before a reward might be potentially the cause, and the assumption is that the network does not know which synapse (or set of synapses) are responsible for the reward (thus the whole network is modulated). 

The simulation of this section is a basic evaluation of a weight updating process. The term $m(t) \cdot E(t)$, which affects Eqs.\,(\ref{eq.m}) and (\ref{eq.wUpdateNew}), and expresses a credit assignment, is pre\hyp determined according to different stochastic regimes. The purpose is to evaluate the difference between single\hyp weight\hyp component and two weight\hyp component dynamics illustrated by Eqs.\,(\ref{eq.W}) and (\ref{eq.ltwNewUpdate}), independently of specific reward-learning plasticity rule. 

The value of a weight $\sigma$ is monitored each time an update occurs. Let us assume arbitrarily that a correlation across $\sigma$ and a following unrelated reward occurs coincidentally every five minutes. Three cases are considered. In phase one, the weight is active coincidentally before reward episodes (i.e. no correlation with the reward). For this reason, modulation causes sometimes increments and sometimes decrements. Such setting represents algorithms that do not have an ``unsupervised bias'', e.g.\,\cite{urbanczikSenn2009,fremauxSprekelerGerstner2010}, which guarantee that the reward maximization function has a null gradient if the weight does not cause a reward. To reproduce this condition here, the stochastic updates in phase 1 have an expected value of zero. In a second phase, weight updates cease to occur, representing the fact that the weight $\sigma$ is never active before rewards (no ambiguity in the information flow). In a third phase, the weight $\sigma$ is active before rewards more often than not, i.e.\,it is now mildly correlated to reward episodes, but in a highly stochastic regime. 

\begin{figure}
\begin{center}
\includegraphics[width = 0.75\columnwidth]{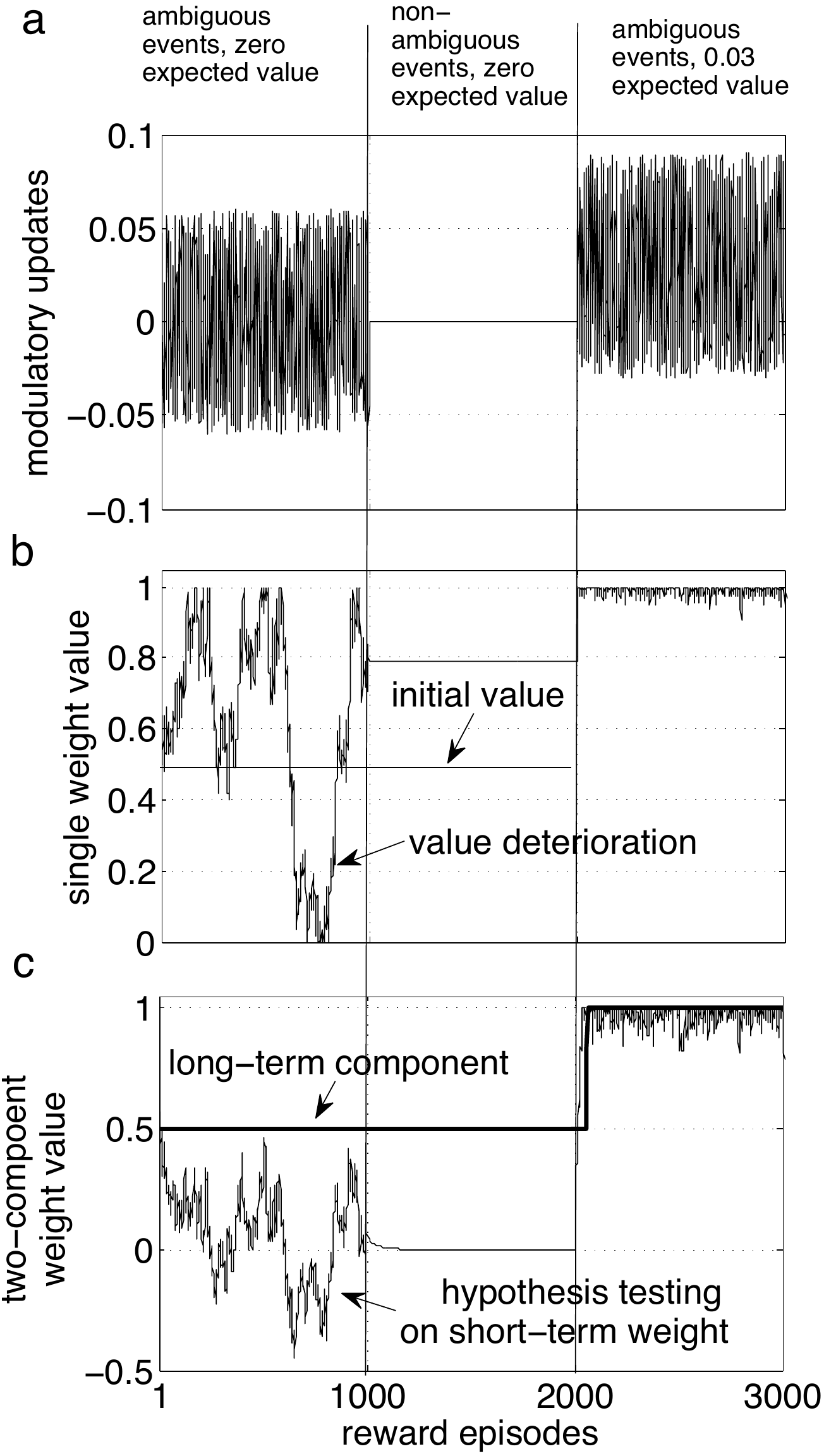}
\caption{\small Illustration of one versus two weight component dynamics with ambiguous updates due to distal rewards. (a) Random values of weight updates illustrate three cases: from 1 to 1000 the weight is not related to a reward, the updates have an expected value of zero. From 1001 to 2000, the weight is never active when rewards occur, there are no updates. From 2001 to 3000, random updates have non-zero mean, i.e. the activity of the weight is correlated to a reward signal. (b) Changes in the weight $\sigma$ when $\sigma$ is composed of one single component. The initial value of 0.5 is progressively forgotten. (c) Changes in the weight of $\sigma$ when $\sigma$ is composed of two components. The initial value of 0.5 is preserved by the long\hyp term component, while the short\hyp term component acts as a monitor for correlations with the reward signal (correctly detected after step 2000).}
\label{fig.drift}
\end{center}
\end{figure} 

Fig.\,\ref{fig.drift}a illustrates weight updates that were randomly generated and draw from the distributions U(-0.06,0.06) for the reward episodes 1 to 1000, U(0,0) for the reward episodes from 1001 to 2000, and U(-0.03,0.09) for the reward episodes 2001 to 3000. The distribution in the first 1000 reward episodes represents a random signal with an expected value of zero, i.e.\,the weight $\sigma$ is not associated with the reward. Figs.\,\ref{fig.drift}bc show respectively the behaviors of a single\hyp weight\hyp component rule and of a two\hyp weight\hyp component rule with weight decay on the short\hyp term component. In the single\hyp weight\hyp component case (Fig.\,\ref{fig.drift}b), despite the updates have an expected value of zero, the weight loses its original value of $0.5$. The initial value of $0.5$ is chosen arbitrarily to be in between $0$ and $1$ to observe both positive and negative variations from its original value. The forgetting of the original value of $\sigma$ is logical because even if the expected value of the updates is zero, there is no mechanism to ``remember'' its initial value. The weight undergoes a random walk, or diffusion, that leads to information loss. The example in Fig.\,\ref{fig.drift}b shows that the weight change is not negligible, ranging from $0$ to saturation. Note that the rate of change, and the difference between the original value and the final value in this example is only illustrative. In a neural network, updates are a function of more variables including the strength of the synapse itself and the neural activity. However, the current example captures an important aspects of learning with delayed rewards: regardless of the plasticity rule, coincidental events in a neural network may lead to unwanted changes. The example is useful to show that a plasticity rule with a single weight-component, even if not affected by the ``unsupervised bias'', disrupts existing weights that are not related to rewards but are active before rewards. Fig.\,\ref{fig.drift}c instead shows that a two\hyp weight\hyp component rule preserves its long\hyp term component, while the short\hyp term component is affected by the random updates. However, due to its decay, the short\hyp term component tends to return to low values if the updates have limited amplitude and an expected value of zero. If rewards and activity across $\sigma$ never occur together (reward episodes from 1001 to 2000), there is no ambiguity and $\sigma$ is clearly not related to rewards: the single\hyp weight\hyp component rule maintains the value of $\sigma$, while the two\hyp weight\hyp component rule has a decay to zero of the short\hyp term component. Finally, in the phase from reward episode 2001 to 3000, the updates have a positive average sign, but are highly stochastic: both rules bring the weight to its saturation value 1. In particular, the two\hyp weight\hyp component rule brings the long\hyp term component to saturation as a consequence of the short\hyp term component being above the threshold level.  

This simple computational example, which does not yet involve a neural model, shows that distal reward learning with a single\hyp weight component leads to weight deterioration of currently non\hyp reward\hyp inducing weights. A two\hyp weight\hyp component rule instead has the potential of preserving the values of weights in the long\hyp term component, while simultaneously monitoring the correlation to reward signals by means of the short\hyp term component. The principle illustrated in this section is used by HTP on a neural model with the results presented in the following sections. 

\subsection{Learning without forgetting}

Three different learning scenarios are devised to test the neural learning with the network in Fig.\,\ref{fig.net}. Each learning scenario lasts 24 h of simulated time and rewards 10 particular stimulus-action pairs (out of a total of 9000 pairs). A scenario may be seen as a learning task composed of 10 subtasks (i.e.\,10 stimulus\hyp action pairs). The aim is to show the capability of the plasticity rule to learn and memorize stimulus\hyp action pairs across multiple scenarios. Note that the plasticity rule is expected to bring to a maximum value all synapses that represent reward-inducing pairs (Fig.\,\ref{fig.net}).

The network was simulated in scenario 1 (for 24 h), then in scenario 2 (additional 24 h), and finally in scenario 3 (again 24h). During the first 24 h (scenario 1), the rewarding input\hyp output pairs are chosen arbitrarily to be those with indices $(i,i)$ with  $1 \leq i \leq 10$. When a rewarding pair occurs, the input $r(t)$ (normally 0) is set to $0.5\pm 0.25$ at time $t + \varphi$ with $\varphi$ drawn from a uniform distribution $U(1,4)$. $\varphi$ represents the delay of the reward. With this setting, not only is a reward occurring with a random variable delay, but its intensity is also random, making the solution of the problem even more challenging. In the second scenario, the rewarding input\hyp output pairs are $(i,i-5)$ with $11 \leq i \leq 20$. No reward is delivered when other stimulus\hyp action pairs are active. A third scenario has again different rewarding pairs as summarized in Table \ref{tab.scenarios}. The arbitrary stimulus-action rewarding pairs were chosen to be easily seen on the weight matrix as diagonal patterns. While stimuli in the interval 31 to 300 occur in all scenarios, stimuli 1 to 10 occur only scenario 1, stimuli 11 to 20 in scenario 2 and stimuli 21 to 30 in scenario 3. This setting is meant to represents the fact that the stimuli that characterize rewards  in one scenario are not present in other scenarios, otherwise all scenarios would be effectively just one. While in theory it would be possible to learn all relationships simultaneously, such a division in tasks (or scenarios) is intended to test learning, memory and forgetting when performing different tasks at different times. It is also possible to interpret a task as a focused learning session in which only a subset of all relationships are observed.  

\begin{table}
\begin{tabular}{|p{1cm}|p{3cm}|c|}
\hline
Scenario & Rewarding stimulus-action pairs & Perceived stimuli\\
\hline
1 & (1,1);(2,2)...(10,10) & 1 to 10 and 31 to 300\\
2 & (11,6);(12,7)...(20,15) & 11 to 20 and 31 to 300\\
3 & (21,1);(22,2)...(30,10) & 21 to 300\\
\hline
\end{tabular}
\caption{Summary of learning scenarios, rewarding stimulus\hyp action pairs, and pool of perceived stimuli. When one of the listed stimulus\hyp action pair occurs, a stochastic reward drawn from the uniform distribution $U(0.25,0.75)$ is delivered with a variable delay between $1$ and $4$ s.}
\label{tab.scenarios}
\end{table} 
\begin{figure*}
\begin{centering}
\includegraphics[width = 0.7\textwidth]{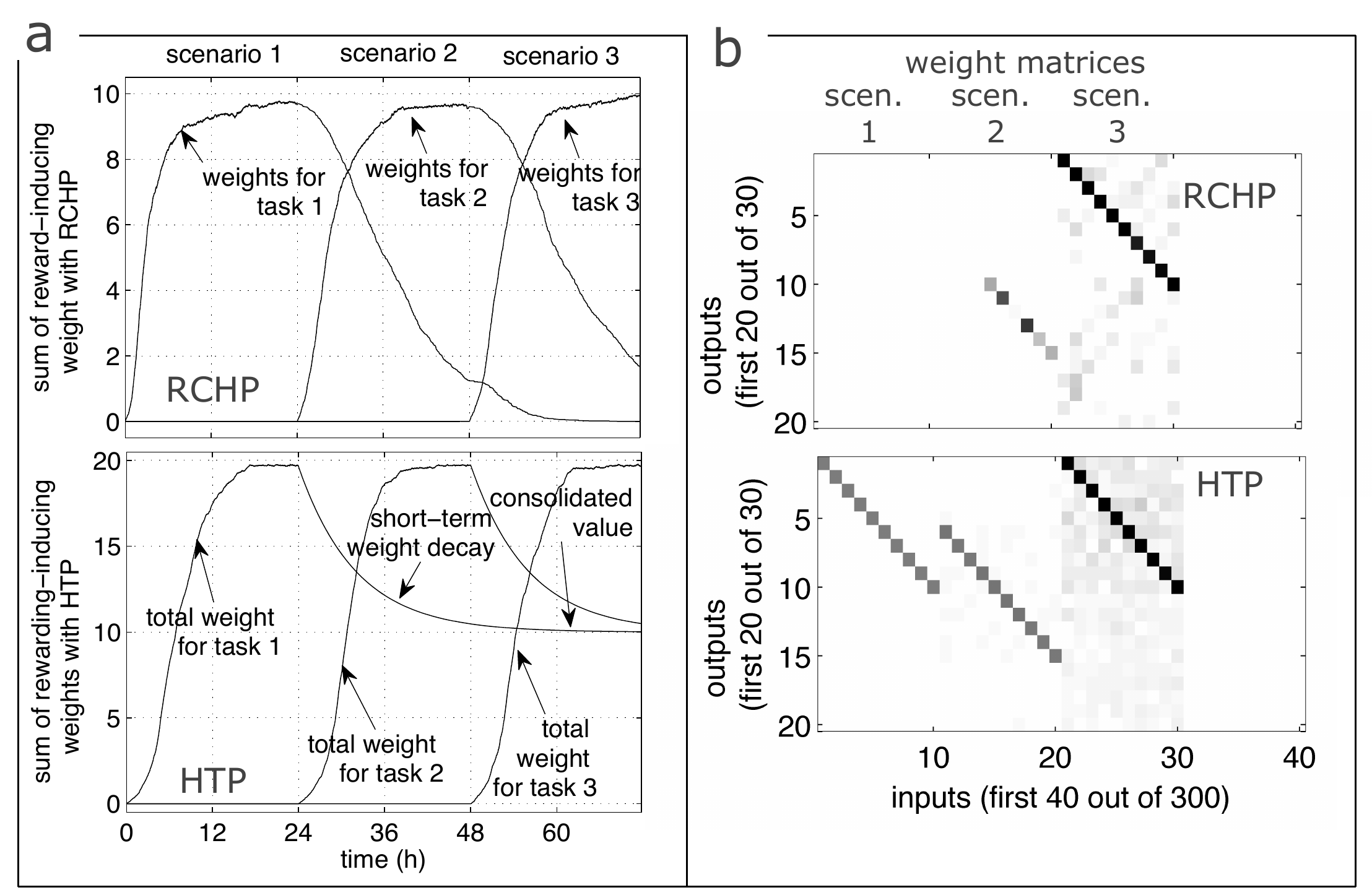}
\caption{\small Learning in three consecutive scenarios. (a) The cumulative total weight of the 10 rewarding synapses (averaged over 10 independent simulations) is shown during the 72 h learning with both RCHP (top graph) and HTP (bottom graph). In the first scenario (first 24 h), the learning leads to a correct potentiation of most reward-inducing synapses. The learning in a second and third scenario with RCHP causes a progressive dismantling of the weights that were reinforced before. HTP identifies consistently all reward-inducing synapses, and does not forget the knowledge of scenario 1 while learning scenario 2 and 3. The partial decay of weights with HTP is due to the short\hyp term component. (b) Partial view of the weight matrix at the end of the 72 h simulation. The view is partial because the overall matrix is 30 by 300: the image only shows the part of the matrix where weights representing relationships are learned. The color represents the strength of the weights, from white (minimum value) to black (maximum value). The high synaptic weights observable under RCHP are those related to scenario 3, because scenarios 1 and 2 are forgotten. The weight matrix with HTP has clearly identified and maintained the 10 rewarding pairs in each scenario.}
\label{fig.f4-3scen}
\end{centering}
\end{figure*} 

Fig.\,\ref{fig.f4-3scen}a shows the cumulative weights of the reward-causing synapses throughout the 72 h of simulation, i.e.\,scenario 1, followed by scenario 2, followed by scenario 3. RCHP, while learning in the second scenario, causes a progressive forgetting of the knowledge acquired during the first scenario. HTP, when learning in scenario 2, also experiences a partial decay of the weights learned during scenario 1. The partial decay corresponds to the short\hyp term weight components. While learning in scenario 2, which represents effectively a different environment, the stimuli of scenario 1 are absent, and the short\hyp term components of the relative weights decay to zero. In other words, while learning in scenario 2, the hypotheses on stimulus\hyp action pairs in scenario 1 are forgotten, as in fact hypotheses cannot be tested in the absence of stimuli. However, the long\hyp term components, which were consolidated during learning in scenario 1, are not forgotten while learning in scenario 2. Similarly it happens in scenario 3. These dynamics lead to a final state of the networks shown in Fig.\,\ref{fig.f4-3scen}b. The weight matrices show that, at the end of the 72 h simulation, RCHP encodes in the weights the reward\hyp inducing synapses of scenario 3, but has forgotten the reward\hyp inducing synapses of scenario 1 and 2. Even with a slower learning rate, RCHP would deteriorate weights that are not currently causing a reward because coincidental correlations and decorrelations alter all weights in the network. In contrast, the long-term component in HTP is immune to single correlation or decorrelation episodes, and thus it is preserved. 

Learning without forgetting with distal rewards is for the first time modeled in the current study by introducing the assumption that established relationships in the environments, i.e.\,long\hyp term weights, are stable and no longer subject to hypothesis evaluation. 

\subsection{The benefit of memory and the preservation of weights}

The distinction between short and long\hyp term weight components was shown in the previous simulation to maintain the memory of scenario 1 while learning in scenario 2, and both scenarios 1 and 2 while learning in scenario 3. 
One question is whether the preservation of long\hyp term weights is effectively useful when revisiting a previously learned scenario. A second fundamental question in this study is whether all weights, reward\hyp inducing and non\hyp reward\hyp inducing, are effectively preserved. To investigate these two points, the simulation was continued for additional 24 h in which the previously seen scenario 1 was revisited.  

\begin{figure}
\includegraphics[width = 0.7\columnwidth]{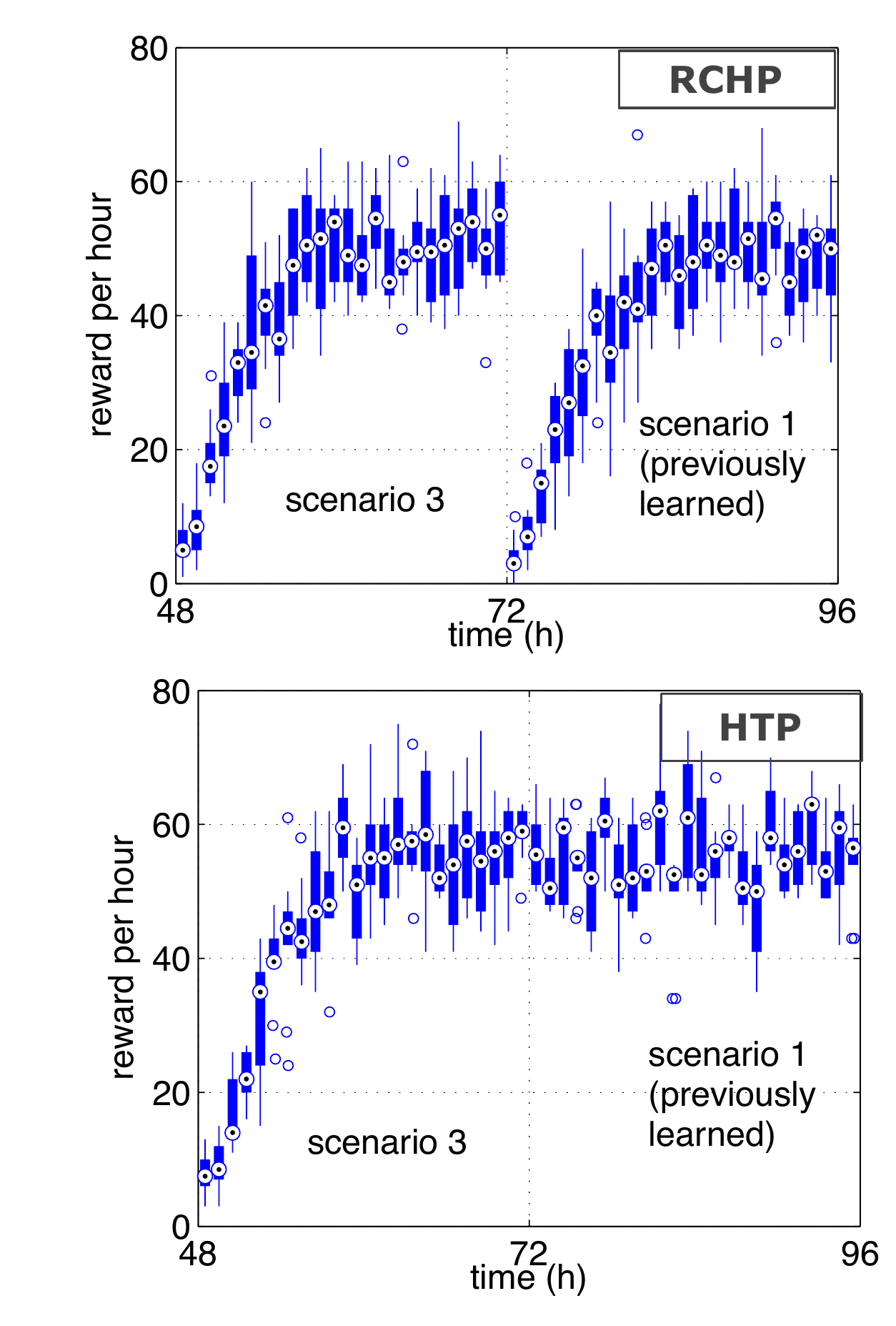}
\begin{centering}
\caption{\small Amount of reward per hour (box plot statistics over 10 independent trials). RCHP, when revisiting scenario 1, needs to relearn the reward-inducing synapses: those weights were reinforced initially (simulation time 0-24 h), but later at time 72 h, those weights, which where not rewarded, deteriorated and dropped to low values. Although relearning demonstrates the capability of solving the distal reward problem, the network with HTP instead demonstrates that knowledge is preserved and reward rates are immediately high when revisiting scenario 1.}
\label{fig.f6-rew}
\end{centering}
\end{figure} 

The utility of memory is shown with the rate of reward per hour as shown in Fig.\,\ref{fig.f6-rew}. RCHP performs poorly when scenario 1 is revisited: it re-learns it as if it had never seen it before. HTP instead performs immediately well because the network remembers the stimulus-response pairs in scenario 1 that were learned 72 hours before. Under the present conditions, long\hyp term weights are preserved indefinitely, so that further learning scenarios can be presented to the network without compromising the knowledge acquired previously.

\begin{figure}
\begin{centering}
\includegraphics[width = 0.35\textwidth]{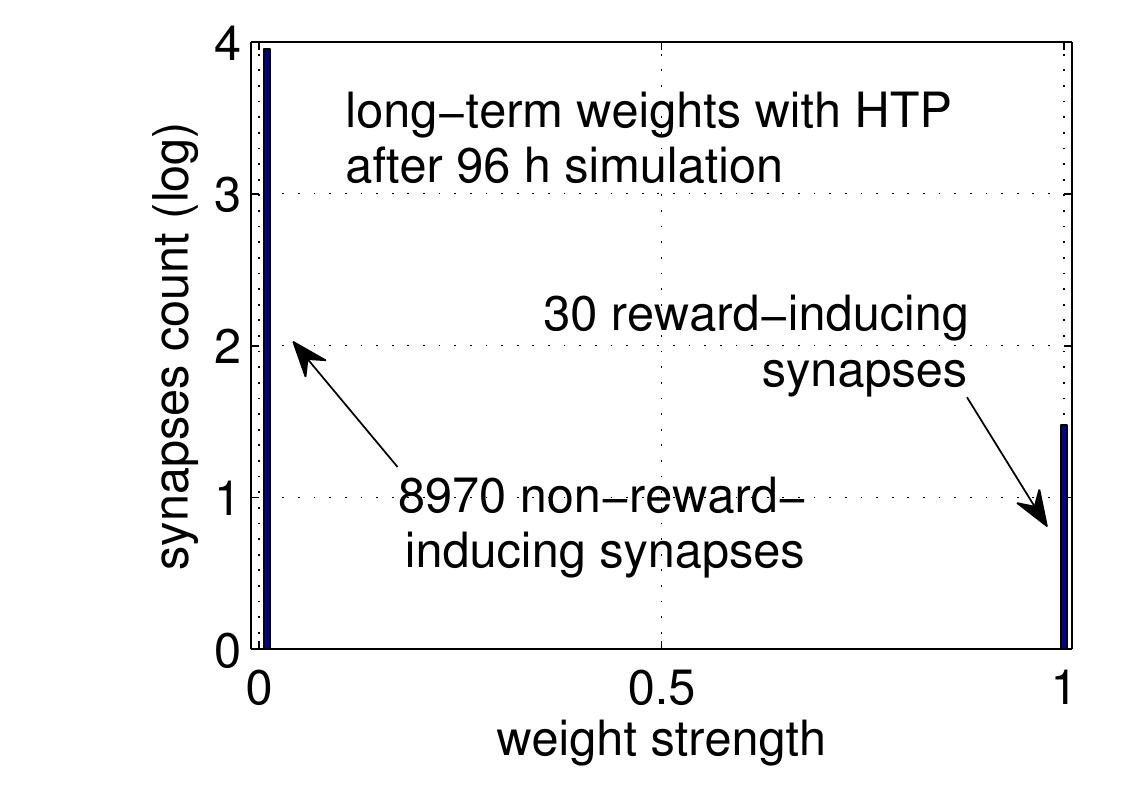}
\caption{Histogram of the long\hyp term weights with HTP after the 96 h of simulation, i.e.\,after performing in scenarios 1, 2, 3 and then 1 again. The long\hyp term components of the weights represent the reward\hyp inducing synapses (an arbitrary set of 30 synapses). All the 8970 non\hyp reward\hyp inducing synapses remain with null weight. This means that the network has not changed any of the weights that are not reward\hyp related. On the other hand, all 30 reward-inducing synapses are identified and correctly consolidated in long-term memory.}
\label{fig.histLT}
\end{centering}
\end{figure}

Eq.\,(\ref{eq.ltwNewUpdate}) allows long\hyp term weights to increase, but not to decrease. Therefore, the analysis of weight changes is simplified in the sense that null long\hyp term components at the end of the run are guaranteed not to have experienced any change. Fig.\,\ref{fig.histLT} shows the histogram of the long\hyp term synaptic weights after 96 h of simulation with HTP. After hundreds of thousand of stimulus-action pairs, and thousands of reward episodes, none of the 8970 synapses representing non\hyp reward\hyp inducing stimulus\hyp action pairs was modified. Those weights were initially set to zero, and remained so, demonstrating that the stable configuration of the network was not altered during distal reward learning. This fact is remarkable considering that the probability of activation of all 9000 pairs is initially equal, and that many disturbing stimuli and non\hyp rewarding pairs are active each time a delayed reward is delivered. This accuracy and robustness is a direct consequence of the hypothesis testing dynamics in the current model: short\hyp term weights can reach high values, and therefore can be consolidated in long\hyp term weights, only if correlations across those weights are consistently followed by a reward. If not, the long\hyp term component of weights is immune to deterioration and preserves its original value.

\subsection{Improved disambiguating capabilities and consequences for learning speed and reliability}

An interesting aspect of HTP is that the change of short\hyp term weights also affects the overall weight W in Eq.\,(\ref{eq.W}). Thus, an update of $w_{st}$ also changes (although only in the short term) how input signals affect output neurons, thereby also changing the decision policy of the network. Initially, when all weights are low, actions are mainly determined by noise in the neural system (introduced in Eq.\,(\ref{eq.neuron})). The noise provides an unbiased mechanism to explore the stimulus-action space. As more rewards are delivered, and hypotheses are formed (i.e.\,weights increase), exploration is biased towards stimulus-action pairs that were active in the past before reward delivery. Those pairs include also non\hyp reward\hyp inducing pairs that were active coincidentally, but they certainly include the reward\hyp triggering ones. Such dynamics have two consequences according to whether a reward occurs or not. In the case a reward occurs again, the network will strengthen even more particular weights which are indeed even more likely to be associated with rewards. To the observer, who does not know at which point short\hyp term weights are consolidated in long\hyp term, i.e. when hypotheses are consolidated in certainties, the network acts as if it knows already, although in reality is guessing (and guessing correctly). By doing so, the network actively explores certain stimulus-action pairs that appear ``promising'' given the past evidence. 

The active exploration of a subset of stimulus-action pairs is particularly effective also when a reward fails to occur, i.e.\,when one hypothesis is false. The negative baseline modulation (term $b$ in Eq.\,(\ref{eq.m})) implies that stimulus-action pairs with high eligibility traces (i.e. that were active in the recent past) but are not followed by rewards decrease their short\hyp term weight components. In a way, the network acts as if trying out  potentially reward-causing pairs (pairs whose weight was increased previously), and when rewards do not occur, drops their values, effectively updating the belief by lowering the short\hyp term components of those weights. %For this reason, the plasticity rule presented in this study implements effectively an hypothesis testing mechanism that increases the weights when rewards occur and decreases the weights when rewards fail to occur. If the weight of a stimulus-action pair was decreased in the past because no reward followed, such a pair becomes less likely to be activated when the same stimuli occur.  

\begin{figure}
\begin{centering}
\includegraphics[width = 0.95\columnwidth]{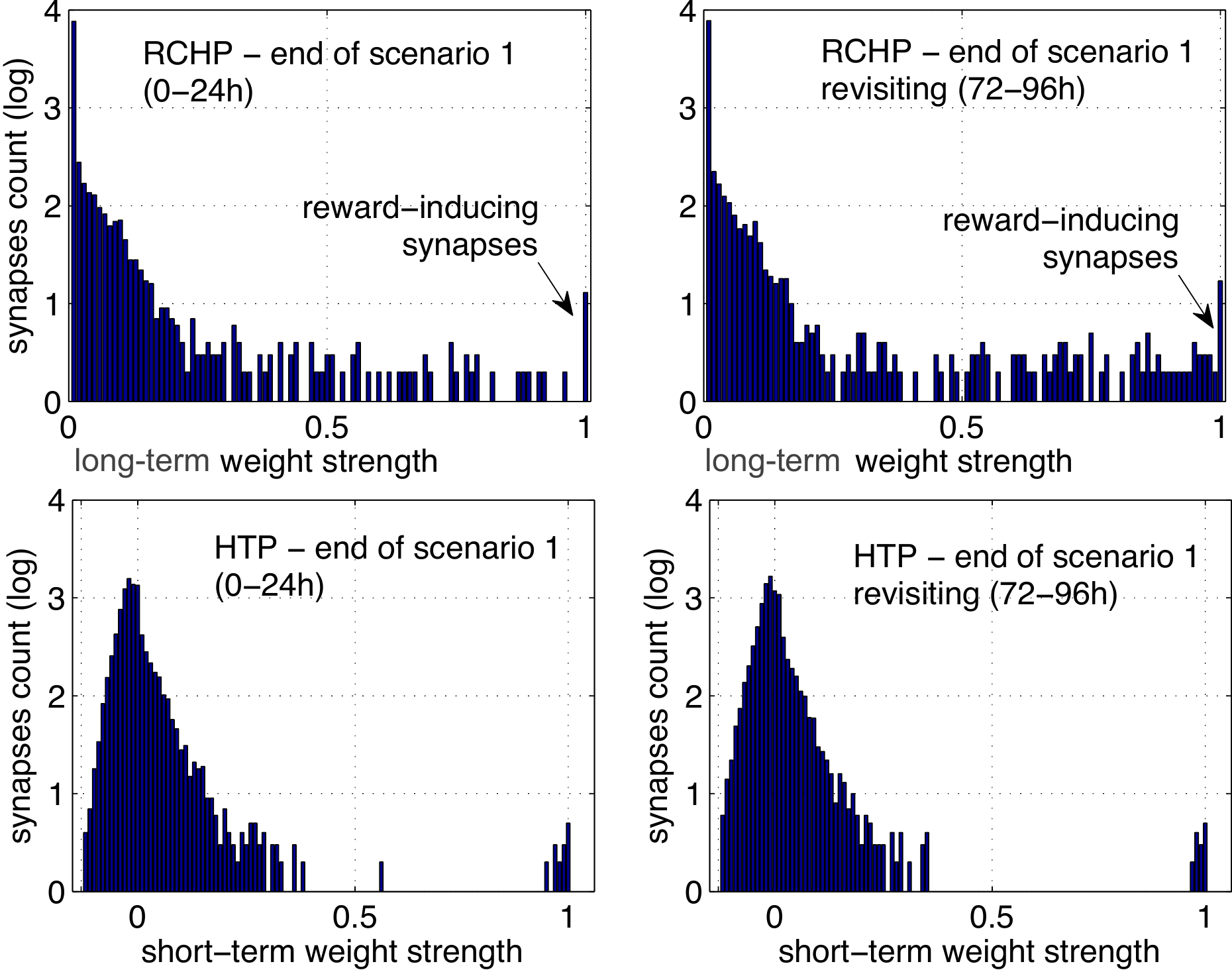}
\caption{Histograms of the weight distribution after learning (long-term total weight for RCHP and short\hyp term for HTP). RCHP (upper graphs) does not appear to separate well the reward-inducing synapses from the others. In particular, in the last phase of the simulation (h 72-96, upper right graph), many synapses reach high values. HTP instead (lower graphs) separates distinctly the short\hyp term components of reward\hyp inducing synapses from the others. At the end of the last simulation (96 h, lower right graph), the separation remains as large as it was at 24 h, indicating that such a weight distribution is stable.}
\label{fig.f3}
\end{centering}
\end{figure}

What are the consequences of these dynamics? An answer is provided by the weight distribution at the end of learning. The histograms in Fig.\,\ref{fig.f3} show that, in contrast to the single-weight rule (upper histograms), HTP separates clearly the reward\hyp inducing synapses from the others (lower histograms). Such a clear separation is then exploited by HTP by means of the threshold $\Psi$ to consolidate reward\hyp inducing weights. The clear separation also provides an insight onto why HTP appeared so reliable in the present experiments. In contrast, RCHP alone cannot separate synapses very distinctly. Such a lack of separation between reward\hyp inducing and non\hyp reward\hyp inducing weights can also be observed in \cite{izhikevich2007,obrienSrinivasa2013NC}. Large synapses in the run with RCHP represent, like for HTP, hypotheses on input\hyp output\hyp reward temporal patterns. However, weights representing false hypotheses are not easily depressed under RCHP or R\hyp STDP that rely only on decorrelations to depress weights. In fact, a large weight causes that synapse to correlate even more frequently, biasing the exploration policy, and making the probability of such an event to occur coincidentally before a reward even higher. Such a limitation in the models in \cite{izhikevich2007,florian2007,obrienSrinivasa2013NC,soltoggioSteilNeuralComputation2013} is removed in the current model that instead explicitly depresses synapses that are active but fail to trigger rewards. Note that HTP pushes also some short\hyp term weights below zero. Those are synapses that were active often but no reward followed. In turn, these lower weights are unlikely to trigger actions. % Such dynamics illustrate the capability of the rule of expressing the probability of synapses to cause a reward: low values or negative valued synapses are very unlikely to be involved in reward\hyp triggering behaviors.

\begin{figure}
\begin{centering}
\includegraphics[width = 0.8\columnwidth]{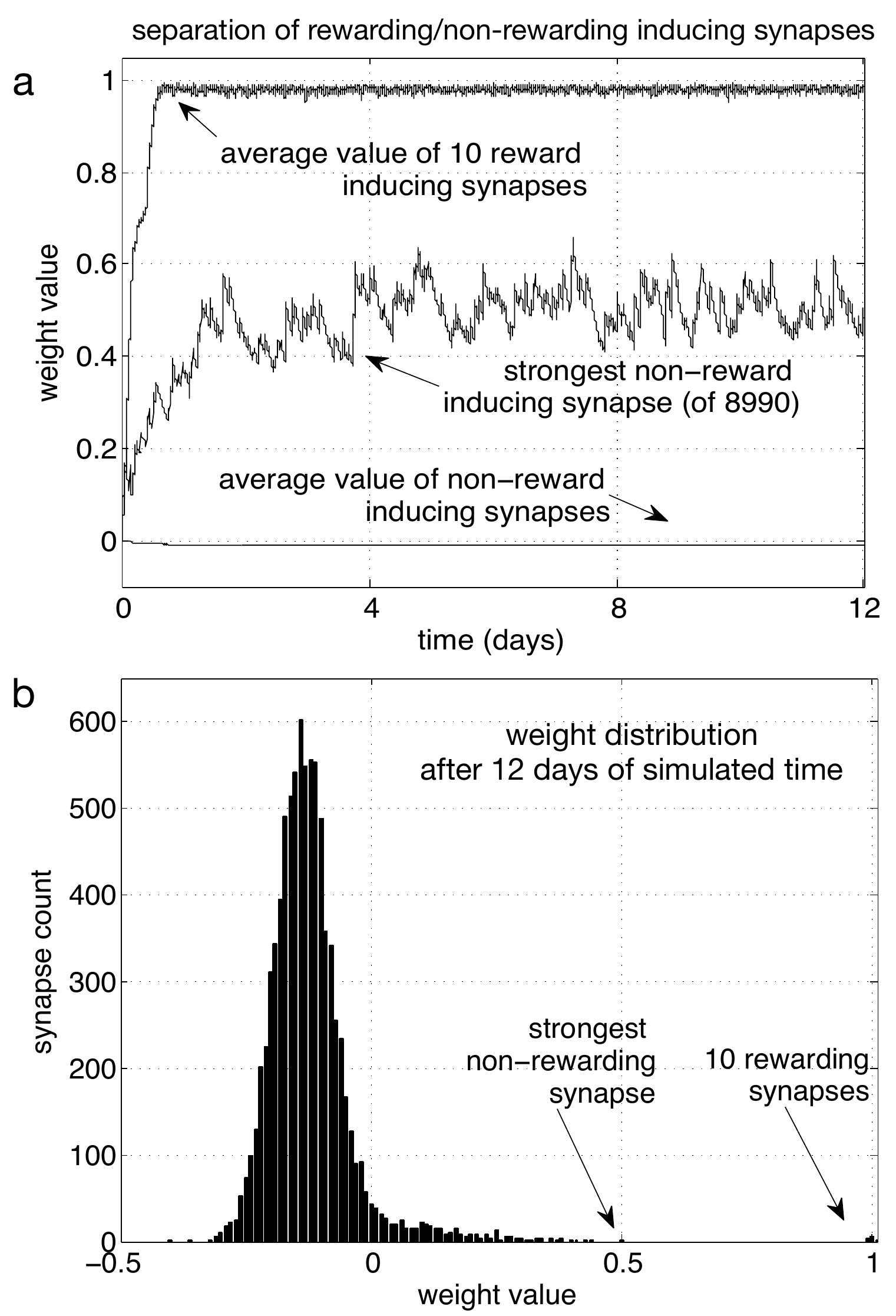}
\caption{Separation of reward\hyp inducing and non\hyp reward\hyp inducing synapses monitored during a long simulation for assessing stability. (a) Scenario 1 is simulated for 12 days (288 simulated hours). The plot shows the average value of the 10 reward\hyp inducing synapses, the strongest individual synapse among the other  $8\,990$ non\hyp reward\hyp inducing synapses, and the average value of all non\hyp reward\hyp inducing synapses. (b) At the end of the 12 days of simulation, the distribution of all weights is shown in the histogram. The number of non\hyp reward\hyp inducing synapses that is potentiated beyond the value $0.1$ is only $2.1$\% of the total.}
\label{fig.monitor}
\end{centering}
\end{figure}

Fig.\,\ref{fig.f3} shows the weight distribution and the separation between reward\hyp inducing and non\hyp reward\hyp inducing synapses at the end of a $4$-day simulated time. One might ask whether this separation and distribution is stable throughout the simulation and over a longer simulation time. One additional experiment was performed by running the learning process in scenario 1 for 12 days of simulated time, i.e.\,an extended amount of time beyond the initial few hours of learning. Fig.\,\ref{fig.monitor}a 
shows the average value of the reward\hyp inducing synapses, the average value of non\hyp reward\hyp inducing synapses and the strongest synapse among the non\hyp reward\hyp inducing ones. The consistent separation in weight between synapses that do or do not induce a delayed reward indicates that the value of $\Psi$, set to $0.95$ in all experiments of this study, is not a critical parameter. If the plasticity rule is capable of separating clearly the reward\hyp inducing synapses from the non\hyp reward\hyp inducing synapses, the parameter $\Psi$ can be set to any high value that is unlikely to be reached by non\hyp reward\hyp inducing synapses. Fig.\,\ref{fig.monitor}b plots the histogram of weight distribution at the end of the simulation (after 12 days of simulated time). The histogram shows clearly that although the strongest non\hyp reward\hyp inducing synapses throughout the run oscillates approximately around $0.5$, the percentage of non\hyp reward\hyp inducing synapses that are potentiated is very small (only 2\% of synapses exceed $0.1$ in strength). 

The fact that HTP separates more clearly rewarding from non\hyp rewarding weights has a fundamental consequence on the potential speed of learning. In fact, high learning rates in ambiguous environments are often the cause of erroneous learning. If a stimulus\hyp action pair appears coincidentally a few times before a reward, a fast learning rate will increase the weight of this pair to high values, leading to what can be compared to superstitious learning \citep{skinner1948,ono1987superstitious}. However, if HTP, for the reasons explained above, is capable of better separation between reward\hyp inducing and non\hyp reward\hyp inducing weights, and in particular is capable of depressing false hypotheses, the consequence is that HTP can adopt a faster learning rate with a decreased risk of superstitious learning.

This section showed that the hypothesis testing rule can improve the quality of learning by (a) biasing the exploration towards stimulus\hyp action pairs that were active before rewards and (b) avoiding the repetition of stimulus\hyp action pairs that in the past did not lead to a reward. In turn, such dynamics cause a clearer separation between reward\hyp inducing synapses and the others, implementing an efficient and potentially faster mechanism to extract cause\hyp effect relationships in a deceiving environment. 

\subsection{Discovering arbitrary reward patterns}

When multiple stimulus\hyp action pairs cause a reward, three cases may occur: 1) each stimulus and each action may be associated to one and only one reward-inducing pair; 2) one action may be activated by more stimuli to obtain a reward; 3) one stimulus may activate different actions to obtain a reward. The cases 1) and 2) were presented in the previous experiments. The case 3) is particular: if more than one action can be activated to obtain a reward, given a certain stimulus, the network may discover one of those actions, and then exploit such pair without learning which other actions also lead to rewards. These dynamics represent an agent who exploits one rewarding action but performs poor exploration, and therefore fails to discover all possible rewarding actions. However, if exploration is enforced occasionally even during exploitation, in the long term the network may discover all actions that lead to a reward given one particular stimulus. To test the capability of the network in this particular case, two new scenarios are devised to reward all pairs identified by a checker board pattern on the weight matrix in a 6 by 12 rectangle, in which each scenario rewards the network that discovers the connectivity pattern of a single 6 by 6 checker board. Each stimulus in the range 1 to 6 in a first scenario, and 7 to 12 in a second scenario, can trigger three different actions to obtain a reward. The two tasks were performed sequentially and lasted each 48 h of simulated time.

\begin{figure*}
\begin{centering}
\includegraphics[width = 0.7\textwidth]{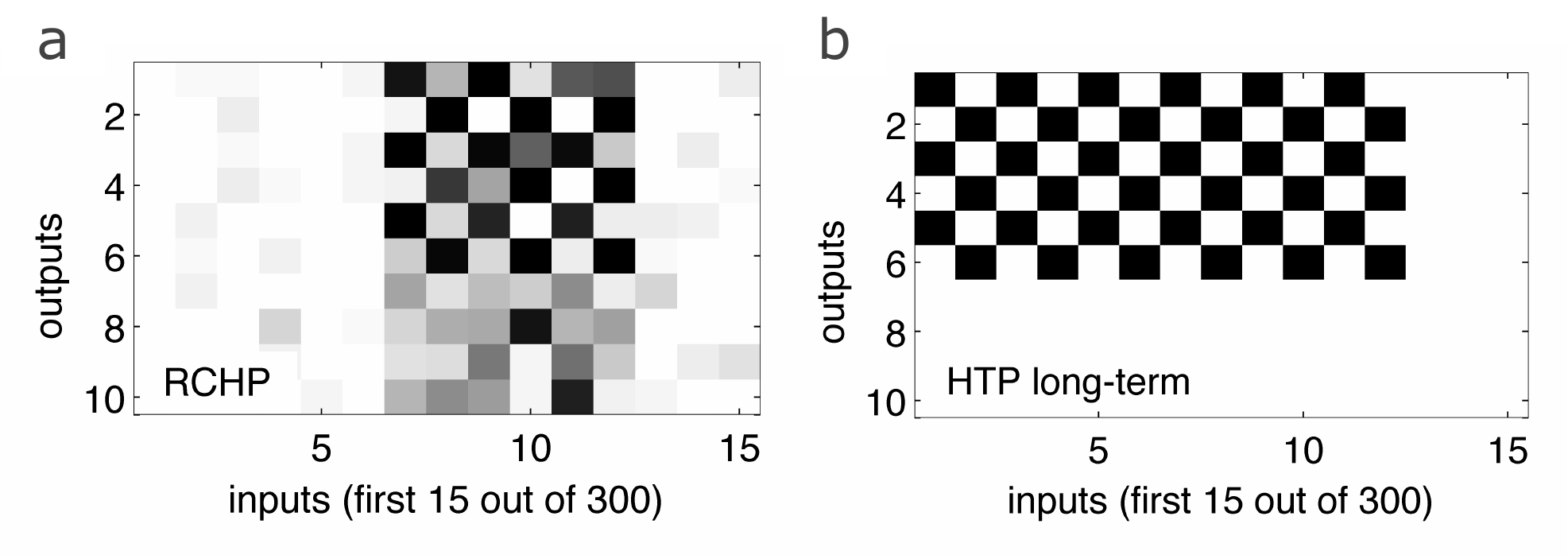}
\caption{Learning arbitrary connectivity patterns: partial view of weight matrices. The color represents the strength of the weights, from white (minimum value) to black (maximum value). (a) RCHP attempts to learn a checker board pattern on 12 inputs and 6 outputs in two consecutive scenarios. After 96 h of simulated time, the rule has discovered an approximation of the pattern for the second task (inputs 7 to 12) but has forgotten the first task. The strengths of the weights do not represent very accurately the reward\hyp inducing pairs (compare with panel b). (b) HTP discovers the exact pattern of connectivity that represents reward conditions in the environment across two scenarios that are learned in sequence.}
\label{fig.lastFig}
\end{centering}
\end{figure*} 
A first preliminary test (data not shown), both with RCHP and HTP, revealed that, unsurprisingly, the network discovers one rewarding action for each stimulus and consistently exploits that action to achieve a reward, thereby failing to discover other rewarding actions. Interestingly, such a behavior might be optimal for a reward maximization policy. Nevertheless, a variation of the experiment was attempted to encourage exploration by reducing the neural gain $\gamma$ in Eq.\,(\ref{eq.neuron}) from $0.5$ to $0.1$. The neural gain expresses the effect of inputs on output neurons: by reducing it, internal noise might occasionally lead to exploration even when a stimulus is known to lead to a reward with a given action. Because exploration is performed occasionally while the network exploits the already discovered reward\hyp inducing pairs, hypotheses are also tested sporadically, and therefore need to remain alive for a longer time. The time constant $\tau_{st}$ of the short\hyp term  weight was set in this particular simulation to 24 h.  For the same reason, the number of actions was limited to 10, i.e.\,only 10 output neurons, so that exploration is performed on a slightly reduced search space.

Fig.\,\ref{fig.lastFig} shows the matrixes of the long\hyp term weights after 96 h of simulated time with RCHP (panel a) and with HTP (panel b). RCHP, as already seen in previous experiments, forgets scenario 1 to learn scenario 2. From the matrix in Fig.\,\ref{fig.lastFig}a it is also evident that RCHP did not increase correctly all weights. Some weights that are non\hyp reward\hyp inducing are nevertheless high. It is remarkable instead that HTP (Fig.\,\ref{fig.lastFig}b) discovers the correct connectivity pattern that not only maximizes the reward, but it also represents all rewarding stimulus-action pairs over the two scenarios. The test shows that HTP remains robust even in conditions in which exploration and exploitation are performed simultaneously. The test demonstrates that if the time\hyp constant of transient weights is sufficiently slow, HTP leads to the discovery of reward\hyp inducing weights even if their exploration is performed sporadically.  
%The previous experiments have shown that the HT plasticity can identify very precisely reward\hyp inducing synapses and separate them from the others. A question is how many different patterns can be stored in a network? This section shows that the HT rule can be used to increase the weights of arbitrary reward patterns across stimulus-action pairs. 

\section{Discussion}

The neural model in this study processes input\hyp output streams characterized by ambiguous stimulus\hyp action\hyp reward relationships. Over many repetitions, it distinguishes between coincidentally and causally related \linebreak events. The flow is ambiguous because the observation of one single reward does not allow for the unique identification of the stimulus-action pair that caused it. The level of ambiguity can vary according to the environment and can make the problem more difficult to solve. Ambiguity increases typically with the delay of the reward, with the frequency of the reward, with the simultaneous occurrence of stimuli and actions, and with the paucity of stimulus\hyp action pairs. The parameters in the neural model are set to cope with the level of ambiguity of the given input\hyp output flow. For more ambiguous environments, the learning rate can be reduced, resulting in a slower but more reliable learning. 

HTP proposes a model in which short\hyp term plasticity does not implement only a duration of a memory \citep{sandberg2003working}, but it rather represents the uncertain nature of hypotheses with respect to established facts. Computationally, the advantages of HTP with respect to previous models derive from two features. A first feature is that HTP introduces long\hyp term and short\hyp term components of the weight with different functions: the short\hyp term component tests hypotheses by monitoring correlations; the long\hyp term component consolidates established hypotheses in long\hyp term memory. A second feature is that HTP implements a better exploration: transient weights mean that stimulus-action pairs are \emph{hypotheses} to be tested by means of a targeted exploration of the stimulus\hyp response space. 

Previous models, e.g. \cite{izhikevich2007,friedrichUrbanczikSenn2011,soltoggioSteilNeuralComputation2013}, that solved the distal reward problem with one single weight component, cannot store information in the long term unless those weights are frequently rewarded. In contrast, HTP consolidates established associations to long\hyp term weights. In this respect,  any R-STDP-like learning rule can learn current reward-inducing relationships, but will forget those associations if the network is occupied in learning other tasks.  HTP can build up knowledge incrementally by preserving neural weights that have been established to represent correct associations. HTP is the first rule to model incremental acquisition of knowledge with highly uncertain cause-effect relationships due to delayed rewards.

As opposed to most reward modulated plasticity models, e.g.\,\citep{legensteinChaseSchwartzMaass2010,obrienSrinivasa2013NC}, the current network is modulated with raw reward signals. There is not an external value storing expected rewards for a given stimulus\hyp action pair. Such reward predictors are often additional computational or memory units outside the network that help plasticity to work. The current model instead performs all computation within the network. In effect, expected rewards are computed implicitly, and at the end very accurately, by the synaptic weights themselves. In fact, the synaptic weights, representing an indication of the probability of a future reward, do also implicitly represent the expected reward of a given stimulus\hyp action pair. For example, a synaptic weight that was consolidated in long\hyp term weight represents the high expectation of a future reward. The weight matrix in Fig.\,\ref{fig.f4-3scen}b (bottom matrix) is an accurate predictor of all rewarding pairs (30) across three different scenarios.

The last experiment showed that the novel plasticity rule can perform well under highly explorative regimes. As opposed to rules with a single weight component, HTP is capable of both maintaining strong weights for exploiting reward conditions, and exploring new stimulus\hyp action pairs. By imposing an arbitrary set of reward\hyp inducing pairs, e.g.\,the environmental reward conditions are expressed by a checker board on the weight matrix, the last experiment showed that HTP can use very effectively the memory capacity of the network.

The model can also be seen as a high-level abstraction of memory consolidation \citep{mcGaugh2000,baileyGiustettoHuangHawkinsKandel2000,lamprecht2004structural,dudai2004,mayford2012synapses} under the effect of delayed dopaminergic activity \citep{jay2003}, particularly at the synaptic level as the transition from early-phase to late-phase LTP \citep{lynch2004rev,clopathZieglerVasilakiBsingGerstner2008}. The consolidation process, in particular, expresses a meta\hyp plasticity mechanism \citep{abrahamBear1996,abrahamRobins2005,abraham2008}, with similarities to the cascade model in \cite{fusi2005cascade}, because frequent short\hyp term updates are preconditions for further long\hyp term potentiation \citep{goelet1986long,nguyen1994requirement}. By exploiting synaptic plasticity with two different timescales (short and long\hyp term), the current model also contributes to validating the growing view that multiple timescale plasticity is beneficial in a number of learning and memory models \citep{abbottRegehr2004,fusi2005cascade,fusi2007neural}. The dynamics presented in this study do not reproduce or model biological phenomena \citep{zucker2002short}. Nevertheless, this computational model proposes a link between short\hyp term plasticity and short\hyp term memory, suggesting the utility of fading short-term memories \citep{jonides2008mind}, which may not be a shortcoming of neural systems, but rather a useful computational tool to distinguish between coincidental and reoccurring events. %The hypothesis testing plasticity proposes a novel neural learning mechanism that performs correct disambiguation of confusing events in the world while preserving memory. 

It is interesting to ask which conditions may lead HTP to fail. HTP focuses on and exploits dynamics of previously proposed reward learning rules that aim at separating rewarding pathways from other non\hyp rewarding pathways. Such a separation is not always easy to achieve. For example, in a plot in \cite{izhikevich2007} (Fig. 1d), a histogram of all weights shows that the separation between the rewarding synapse and all other synapses is visible but not large. The original RCHP, as reproduced in this study, may also encounter difficulties in creating clear separations as shown in Fig.\,\ref{fig.f3}. In short, HTP prescribes mechanisms to create a clear separation between reward\hyp inducing and non\hyp reward\hyp inducing synapses: if this cannot be achieved, HTP cannot be used to consolidate long\hyp term weights. This may be the case when the network is flooded with high levels of reward signals. As a general rule, whenever the input\hyp output flow is ambiguous, plasticity rules require time to separate rewarding weights from non\hyp rewarding weights. A fast learning rate is often the cause of failure. Interestingly, a fast learning rate with distal rewards can be imagined as a form of superstitious type of learning, in which conclusions are drawn from few occurrences of rewards \citep{skinner1948,ono1987superstitious}. 
 
If learning rates are small (or similarly if rewards are small in magnitude), would not the decay of transient weights in HTP prevent learning? The answer is that the decay of the transient weights, in this study set to 8h (or 24h for the last experiment), represents the time of one learning scenario. Stimuli, actions and rewards occur in the order of seconds and minutes, so that transient weights do hold their values during a learning phase. In effect, HTP suggests the intuitive notion that learning sessions may need to have a minimum duration or intensity of reward to be effective in the long term. Interestingly, experiments in human learning such as that described in \cite{hamilton1998cortical}, seem to suggest that learning modifies synapses initially only in their short\hyp term components, which decay within days if learning is suspended. A long lasting modification was registered only after months of training \citep{hamilton1998cortical}. An intriguing possibility is that the consolidation of weights does not require months only because of biological limitations (e.g.\, growth of new synapses): the present model suggests that consolidation may require time in order to extract consistent and invariable relationships. So if short\hyp term changes are consistently occurring across the same pathways every week for many weeks, long\hyp term changes will also take place.

The model shows how neural structures may be preserved when learning. From this perspective, it emerges that the  mechanism for learning is the same that preserves memory, effectively highlighting a strong coupling of learning and memory as it also suggested in biology \citep{bouton1994conditioning}. It is nevertheless important to point out that the evidence of associative learning in animals \citep{grossberg1971dynamics,bouton2004memory} depicts yet more complex dynamics that are not captured by current models.

Despite its simplified dynamics with respect to biological systems, the neural learning described by HTP offers a new tool to study learning and cognition both in animals and in neural artificial agents or neuro\hyp robots \citep{krichmarRoehrbein2013}. The proposed dynamics allow for biological and robotics modelling of extended and realistic learning scenarios which were previously too complex for neural models. Examples are learning in interaction where overlapping stimuli, actions, and highly stochastic feedback occur at uncertain times \citep{soltoggioReinhartLemmeSteil2013}. The acquisition of knowledge with HTP can integrate different tasks and scenarios, thereby opening the possibility of studying integrated cognition in unified neural models. This property may in turn result in models for the acquisition of incrementally complex behaviors at different stages of learning \citep{weng2001autonomous,lungarella2003developmental,asada2009cognitive}. 

In the current model, long\hyp term weights do not decay, i.e. they preserve their values indefinitely. This assumption reflects the fact that, if a certain relationship was established, i.e.\,if it was converted from hypothesis to certainty, it represents a fact in the world. To confirm that, the plot in Fig.\,\ref{fig.monitor}a proved that, with a frequency of 1.5 Hz of the stimuli and a 100 ms sampling time, no wrong connection was consolidated in the extended experiment over 288 h of simulated time. The lack of reversal learning (long\hyp term weights cannot decrease) works in this particular case because the environment and tasks in the current study are static, i.e.\,the stimulus\hyp response pairs that induce rewards do not change. Under such conditions, the learning requires no unlearning. However, environments may be indeed changeable, and the rewarding conditions may change over time. In such cases, one simple extension for adaptation is necessary. Assume that one rewarding pair ceases at one point to cause rewards. HTP will correctly detect the case by depressing the short\hyp term weight, i.e.\,the hypothesis becomes negative. In the current algorithm, depression of short\hyp term weights does not affect long\hyp term weights. However, the consolidation described by Eq.\,(\ref{eq.ltwNewUpdate}) can be complemented by a symmetrical mechanism that depresses long\hyp term weights when hypotheses are negative. With such an extension, the model can perform reversal of learning \citep{van1997hebbian,decoRolls2005cerebralCortex,odoherty2001nature}, thereby removing long\hyp term connections when they do not represent anymore correct relationships in the world. The extension to unlearning is shown in the Appendix 1.

\section{Conclusion}

The proposed model introduces the concept of \emph{hypothesis testing} of cause\hyp effect relationships when learning with delayed rewards. The model describes a conceptual distinction between short\hyp term and long\hyp term plasticity, which is not focused on the duration of a memory, but it is rather related to the confidence with which cause\hyp effect relationships are considered consistent \linebreak \citep{abrahamRobins2005}, and therefore preserved as memory. 

The meta\hyp plasticity rule, named \emph{Hypothesis Testing Plasticity} (HTP), models how cause-effect relationships can be extracted from ambiguous information flows, first by validation and then by consolidation to long\hyp term memory. The short\hyp term dynamics boost exploration and discriminate more clearly true cause\hyp effect relationships in a deceiving environment. The targeted conversion of short\hyp term to long\hyp term weights models the consolidation process of hypotheses in established facts, thereby addressing the plasticity\hyp stability dilemma \citep{abrahamRobins2005}.  HTP suggests new cognitive models of biological and machine learning that explain dynamics of learning in complex and rich environments. This study proposes a theoretical motivation for short\hyp term plasticity, which helps hypothesis testing, or learning in deceiving environments, and the following memorization and consolidation process.

\section*{Acknowledgement}

The author thanks John Bullinaria, William Land, Albert Mukovskiy, Kenichi Narioka, Felix Reinhart, Walter Senn, Kenneth Stanley, and Paul Tonelli for constructive discussions and valuable comments on early drafts of the manuscript. A large part of this work was carried out while the author was with the CoR-Lab at Bielefeld University, funded by the European Community's Seventh Framework Programme FP7/2007\hyp2013, Challenge 2â Cognitive Systems, Interaction, Roboticsâ under grant agreement No 248311 - AMARSi.

\section*{Appendix 1: Unlearning}
\label{app.1}

Unlearning of the long\hyp term components of the weights can be effectively implemented as symmetrical to learning. I.e., when the transient weights are very negative (lower than $-\Psi$), the long-term component of a weight is decreased. This process represents the validation of the hypothesis that a certain stimulus-action pair is not associated with a reward anymore, or that is possibly associated with punishment. In such a case, the neural weight that represents this stimulus\hyp action pair is decreased, and so is the probability of occurring.  The conversion of negative transient weights to decrements of long\hyp term weights, similarly to Eq.\, (\ref{eq.ltwNewUpdate}), can be formally expressed as
\begin{equation}
\dot{w}^{lt}_{ji}(t) = -\rho \cdot H(-w^{st}_{ji}(t) - \Psi)\quad.\label{eq.ltwNewUpdateUnlearning}
\end{equation}
No other changes are required to the algorithm described in the paper. 

\begin{figure}
\begin{center}
\includegraphics[width = 0.7\columnwidth]{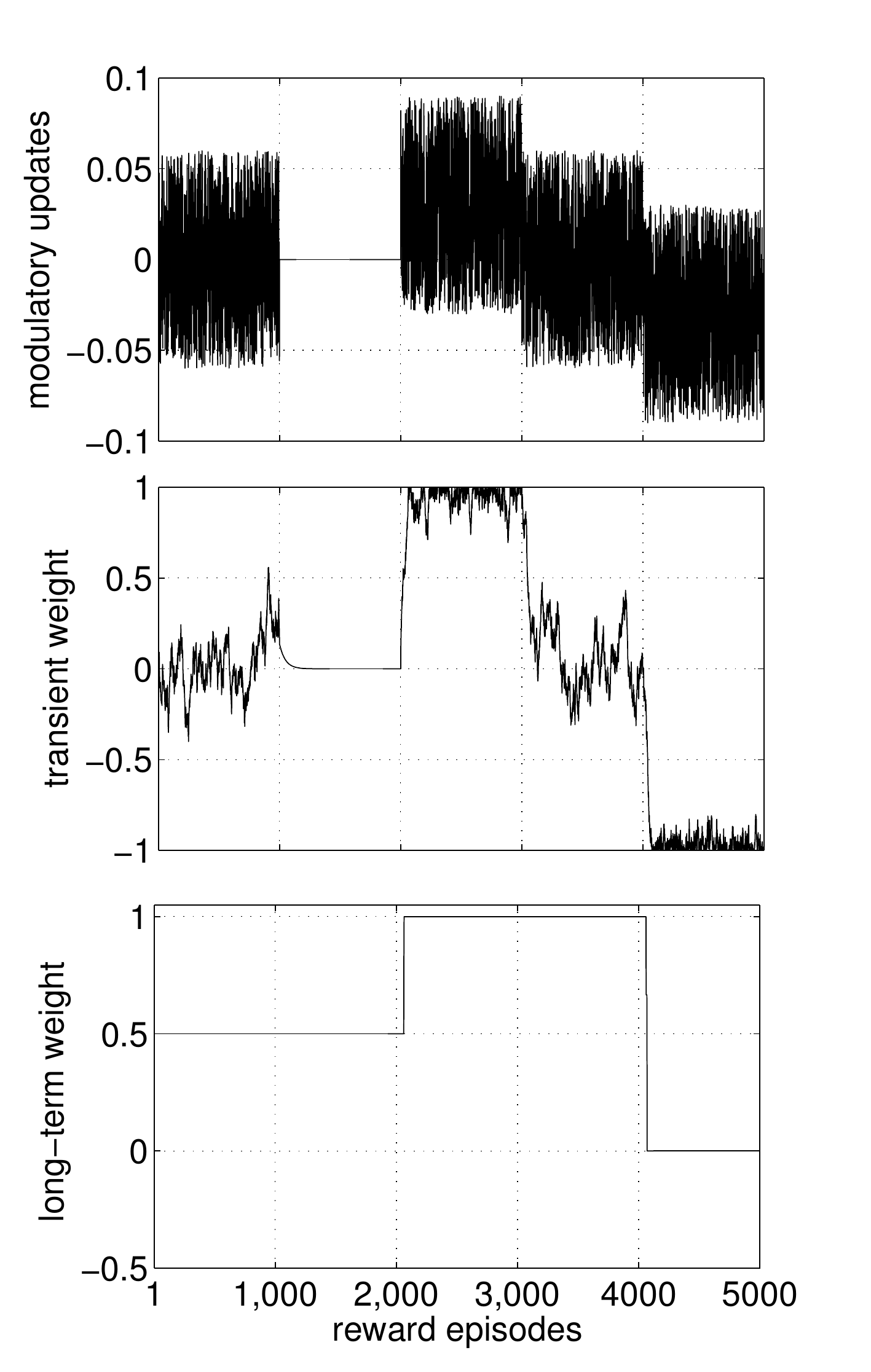}
\caption{Unlearning dynamics. In this experiment, the model presented in the paper was augmented with Eq.\,(\ref{eq.ltwNewUpdateUnlearning}), which decreases long\hyp term weights if the transient weights are lower than $-\Psi$. The stochastic modulatory update (top graph) is set to have a slightly negative average in the last phase (from reward 4001 to 5000). The negative average is detected by the short\hyp term component that becomes negative. The long\hyp term component decreases its value due to Eq.\,(\ref{eq.ltwNewUpdateUnlearning}).}
\label{fig.unlearingOne}
\end{center}
\end{figure}

The case can be illustrated reproducing the preliminary test of Fig.\,\ref{fig.drift}, augmenting it with a phase characterised by a negative average modulation. Fig.\,\ref{fig.unlearingOne} 
shows that, when modulatory updates become negative on average (from reward 4000 to reward 5000), the transient weight detects it by becoming negative. The use of Eq.\,(\ref{eq.ltwNewUpdateUnlearning}) then causes the long\hyp term component to reduce its value, thereby reversing the previous learning. 

Preliminary experiments with unlearning on the complete neural model of this study show that the rate of negative modulation drops drastically as unlearning proceed. In other words, as the network experiences negative modulation, and consequently reduces the frequencies of punishing stimulus\hyp action pairs, it also reduces the rate of unlearning because punishing episodes become sporadic. It appears that unlearning from negative experiences might be slower that learning from positive experiences. Evidence from biology indicates that extinction does not remove completely the previous association \citep{bouton2000learning,bouton2004context}, suggesting that more complex dynamics as those proposed here may regulate this process in animals. 

\section*{Appendix 2: Implementation}

All implementation details are also available as part of the open source Matlab code provided as support material. The code can be used to reproduce the results in this work, or modified to perform further experiments. The source code can be downloaded from\linebreak http://andrea.soltoggio.net/HTP.

\subsection*{Network, inputs, outputs, and rewards}

The network is a feed\hyp forward single layer neural network with 300 inputs, 30 outputs, 9000 weights, and sampling time of 0.1 s.  
Three hundred stimuli are delivered to the network by means of 300 input neurons. Thirty actions are performed by the network by means of 30 output neurons.

The flow of stimuli consists of a random sequence of stimuli each of duration between 1 and 2 s. The probability of 0, 1, 2 or 3 stimuli to be shown to the network simultaneously is described in Table \ref{tab.IOR}.

The agent continuously performs actions chosen form a pool of 30 possibilities. Thirty output neurons may be interpreted as single neurons, or populations. When one action terminates, the output neuron with the highest  activity initiates the next action. Once the response action is started, it lasts a variable time between 1 and 2 s. During this time, the neuron that initiated the action receives a feedback signal I of 0.5. The feedback current enables the output neuron responsible for one action to correlate correctly with the stimulus that is simultaneously active. A feedback signal is also used in \cite{urbanczikSenn2009} to improve the reinforcement learning performance of a neural network. 

The rewarding stimulus\hyp action pairs are $(i,i)$ with $1 \leq i \leq 10$ during scenario 1, $(i,i-5)$ with $11 \leq i \leq 20$ in scenario 2, and $(i,i-20)$ with $21 \leq i \leq 30$ in scenario 3. When a rewarding stimulus-action pair is performed, a reward is delivered to the network with a random delay in the interval [1, 4] s. Given the delay of the reward, and the frequency of stimuli and actions, a number of stimulus\hyp action pairs could be responsible for triggering the reward. The parameters are listed in Table \ref{tab.IOR}.

\begin{table}
\begin{tabular}{l|l}
Parameter&Value\\
\hline
Inputs & 300\\
Outputs & 30\\
Stimulus/input duration & [1, 2] s\\
Max number of active inputs & 3\\
Probability of no stimuli & 1/8\\
Probability of 1 active stimulus & 3/8\\
Probability of 2 active stimuli & 3/8\\
Probability of 3 active stimuli & 1/8\\
Action/output duration & [1, 2] s\\
Rewarding stimulus-action pairs& 30\\
Delay of the reward  & [1, 4] s\\
Nr of scenarios & 3\\
Duration of one learning phase & 24 h\\
\end{tabular}
\caption{Summary of parameters for the input, output and reward signals.}
\label{tab.IOR}
\end{table}

\subsection*{Integration}

The integration of Eqs.\,(\ref{eq.eTupdate}) and (\ref{eq.m}) with a sampling time $\Delta t$ of $100$ ms is implemented step\hyp wise by 
\begin{eqnarray}
E_{ji}(t+\Delta t) &=& E_{ji}(t) \cdot e^{\frac{-\Delta t}{\tau_{E}}} + \mathrm{RCHP}_{ji}(t)\label{eq.c-stepwise}\\
m(t+\Delta t) &=& m(t) \cdot e^{\frac{-\Delta t}{\tau_{m}}} + \lambda\, r(t) + b\label{eq.m-stepwise}\quad.
\end{eqnarray}

The same integration method is used for all leaky integrators used in this study. Given that $r(t)$ is a signal from the environment, it might be a one-step signal as in the present study, which is high for one step when reward is delivered, or any other function representing a reward: in a test of RCHP on the real robot iCub \citep{soltoggioLemmeReinhartSteilFrontiers2013,soltoggioReinhartLemmeSteil2013}, r(t) was determined by the human teacher by pressing skin sensors on the robotÕs arms.

\begin{table}
\begin{tabular}{l|l}
Parameter & Value\\
\hline
Number of neurons & $330$\\
Number of synapses & $9000$\\
Weight range & $[0,1]$\\
Noise on neural transmission ($\xi_{i}(t)$, Eq.\,(\ref{eq.neuron}))& $0.02$ std\\
Sampling time step ($\Delta t$, Eq.\,(\ref{eq.neuron}))& $100$ ms\\
Baseline modulation ($b$ in Eq.\,(\ref{eq.m})) & -0.03 / s\\
Neural gain ($\gamma$, Eq.\,(\ref{eq.neuron})) & $0.5$\\
Short\hyp term learning rate ($\lambda$ in Eqs.\,(\ref{eq.m}) and (\ref{eq.m-stepwise})) & 0.1\\
Time constant of modulation ($\tau_{m}$)& 0.1 s\\
Time constant of traces ($\tau_{E}$) & 4 s\\
\hline
\end{tabular}
\caption{Summary of parameters of the neural model.}
\label{tab.neuron}
\end{table}

\subsection*{Rarely Correlating Hebbian Plasticity}

Rarely Correlating Hebbian Plasticity (RCHP) \linebreak \citep{soltoggioSteilNeuralComputation2013} is a type of Hebbian plasticity that filters out the majority of correlations and produces nonzero values only for a small percentage of synapses. Rate\hyp based neurons can use a Hebbian rule augmented with two thresholds to extract low percentages of correlations and decorrelations. RCHP expressed by Eq.\,(\ref{eq.RCHP}) is simulated with the parameters in Table \ref{tab.RCHP}.
\begin{table}
\begin{tabular}{l|l}
Parameter & Value\\
\hline
Rare correlations ($\mu$ in Eqs.\,(\ref{eq.thetahi}) and (\ref{eq.thetalo})) & $0.1\% / s$\\
Update rate of $\theta$ ($\eta$ in Eqs.\,(\ref{eq.thetahi}) and (\ref{eq.thetalo})) & 0.001 / s\\
$\alpha$ (Eq.\,(\ref{eq.RCHP})) & 1\\
$\beta$ (Eq.\,(\ref{eq.RCHP})) & 1\\
Correlation sliding window (Eq.\,(\ref{eq.rhoCdisc})) & 5 s\\
Short\hyp term time constant ($\tau_{st}$ in Eq.\,(\ref{eq.wUpdateNew})) & 8 h\\
Consolidation rate ($\rho$ in Eq.\,(\ref{eq.ltwNewUpdate})) & $1 / 1800$ s \\
Consolidation threshold ($\Psi$ in Eq.\,(\ref{eq.ltwNewUpdate})) & 0.95\\
\hline
\end{tabular}
\caption{Summary of parameters of the plasticity rules (RCHP and RCHP$^{+}$ plus HTP).}
\label{tab.RCHP}
\end{table}
The rate of correlations can be expressed by a global concentration $\omega_{c}$. This measure represents how much the activity of the network correlates, i.e. how much the network activity is deterministically driven by connections or is instead  noise\hyp driven. The instantaneous matrix of correlations $\mathrm{RCHP}^+$ (i.e.\,the first row in Eq.\,(\ref{eq.RCHP}) computed for all synapses) can be low filtered as
\begin{equation}
\dot{\omega}_{c}(t) = -\frac{\omega_{c}(t)}{\tau_{c}} + \sum_{j = 1}^{300}\sum_{i = 1}^{30} \mathrm{RCHP}_{ji}^+(t)\quad,
\label{eq.rhoCcont}
\end{equation}
to estimate the level of correlations in the recent past, where $j$ is the index of input neurons, and $i$ the index of the output neurons. In the current settings, $\tau_{c}$ was chosen equal to 5 s. Alternatively, a similar measure of recent correlations $\omega_{c}(t)$ can be computed in discrete time over a sliding time window of 5 s summing all correlations $\mathrm{RCHP}^+(t)$
\begin{equation}
\omega_{c}(t) = \Delta t \frac{\sum_{0}^{t-5} \mathrm{RCHP}^+(t)}{5}\quad.
\label{eq.rhoCdisc}
\end{equation}
Similar equations to (\ref{eq.rhoCcont}) and (\ref{eq.rhoCdisc}) are used to estimate decorrelations $\omega_{d}(t)$ from the detected decorrelations $\textrm{RCHP}^-(t)$. The adaptive thresholds $\theta_{hi}$ and $\theta_{lo}$ in Eq.\,(\ref{eq.RCHP}) are estimated as follows. 

%\begin{eqnarray}
%$\theta_{hi}(t+\Delta t) = $\theta_{hi}$ + \kappa \cdot \Delta t \$_{ji}(t+\Delta t) &=& E_{ji}%(t) \cdot e^{\frac{-\Delta t}{\tau_{E}}} + \mathrm{RCHP}_{ji}(t)\label{eq.c-stepwise}\\
%m(t+\Delta t) &=& m(t) \cdot e^{\frac{-\Delta t}{\tau_{m}}} + \lambda\, r(t) + b\label{eq.m-%stepwise}\quad.
%\end{eqnarray}

\begin{equation}
\theta_{hi}(t+\Delta t) = 
\left\{
\begin{array}{ll}
\theta_{hi} + \eta \cdot \Delta t  \,\, &\mathrm{if} \,\,\omega_{c}(t) > 2\mu\\
\theta_{hi} - \eta \cdot \Delta t  \,\, &\mathrm{if} \,\,\omega_{c}(t) < \mu/2\\
\theta_{hi}(t)&\,\, \mathrm{otherwise}
\end{array} \right.
\label{eq.thetahi}
\end{equation}
and
\begin{equation}
\theta_{lo}(t+\Delta t) = 
\left\{
\begin{array}{ll}
\theta_{lo} - \eta \cdot \Delta t  \,\, &\mathrm{if} \,\,\omega_{d}(t) > 2\mu\\
\theta_{lo} + \eta \cdot \Delta t  \,\, &\mathrm{if} \,\,\omega_{d}(t) < \mu/2\\
\theta_{lo}(t)&\mathrm{otherwise}\\
\end{array} \right.
\label{eq.thetalo}
\end{equation}
with $\eta = 0.001$ and $\mu$, the target rate of rare correlations, set to 0.1\%/s. If correlations are lower than half of the target or are greater than twice the target, the thresholds are adapted to the new increased or reduced activity. This heuristic has the purpose of maintaining the thresholds relatively constant and perform adaptation only when correlations are too high or too low for a long period of time. 

%\begin{acknowledgements}
%\end{acknowledgements}

\bibliographystyle{spbasic}      % basic style, author-year citations

%\bibliography{/Users/andrea/work/Ref/filed/theBibNew}

\end{document}